\documentclass[review]{elsarticle}

\usepackage{lineno,hyperref}
\usepackage{xcolor}
\usepackage{mathtools}
\usepackage{algorithm}
\usepackage{algpseudocode}
\usepackage{multirow}
\usepackage{amsfonts,amsmath,amssymb}
\modulolinenumbers[5]
\usepackage{subfig}
\usepackage{comment}

\makeatletter
\renewcommand{\ALG@name}{Pseudocode}
\makeatother

\journal{Journal of \LaTeX\ Templates}

\graphicspath{{images/}}

\begin{document}
	
	\begin{frontmatter}
		
		\title{Multi-Objective Parameter-less Population Pyramid for Solving Industrial Process Planning Problems}
		\tnotetext[mytitlenote]{Fully documented templates are available in the elsarticle package on \href{http://www.ctan.org/tex-archive/macros/latex/contrib/elsarticle}{CTAN}.}

		\author[mymainaddress]{Michal Witold Przewozniczek}
		\cortext[mycorrespondingauthor]{Michal Przewozniczek}
		\ead{michal.przewozniczek@pwr.edu.pl}
		
		\author[mymainaddress]{Piotr Dziurzanski}
		\author[mymainaddress]{Shuai Zhao}
		
		\author[mymainaddress]{Leandro~Soares~Indrusiak}
		
		\address[mymainaddress]{Department of Computer Science, University of York, York, United Kingdom}

		\begin{abstract}
			Evolutionary methods are effective tools for obtaining high-quality results when solving hard practical problems. Linkage learning may increase their effectiveness. One of the state-of-the-art methods that employ linkage learning is the Parameter-less Population Pyramid (P3). P3 is dedicated to solving single-objective problems in discrete domains. Recent research shows that P3 is highly competitive when addressing problems with so-called overlapping blocks, which are typical for practical problems. In this paper, we consider a multi-objective industrial process planning problem that arises from practice and is NP-hard. To handle it, we propose a multi-objective version of P3. The extensive research shows that our proposition outperforms the competing methods for the considered practical problem and typical multi-objective benchmarks.
		\end{abstract}
		
		
	\end{frontmatter}
	
	
	\section{Introduction}
	
	In industry, we often find combinatorial optimisation problems that are non-trivial and NP-hard, which means that in practice, they cannot be solved in polynomial time. Examples of such problems are production process planning or scheduling in single- or multi-objective domains. One of the most common is the Permutation Flow Shop Scheduling Problem (PFSP) \cite{ltgaPopulationSizing,productionSchedulingIeee}. The objective of PFSP is to optimise production quality. A solution in PFSP defines an order in which the production tasks (jobs) are put on the plan by a scheduler. PSFP is considered in single- \cite{ltgaPopulationSizing,productionSchedulingIeee}, multi- \cite{productionSchedulingMO,productionMOrealNumber} and many-objective versions \cite{productionSchedulingManyO}. For some of the problems that emerge from industry optimisation, real numbers are employed to encode a solution \cite{productionMOrealNumber,productionYorkRealNumbers}. For others, sets of discrete values (including binary values) can be used \cite{productionDiscrete,paintsOrig}.\par

	In this paper, we consider a problem of manufacturing process planning in factories producing bulk commodities. Such a process is comprised of manufacturing recipe selection and resource allocation. The main optimisation objective of this case study is to increase production line utilisation and, consequently, to decrease the total production time (makespan) of batch production by executing an appropriate number of recipes producing ordered amounts of commodities. The extension beyond the typical covering problem is that the amount of the commodities produced should be as close to the ordered ones as possible (i.e., the surpluses should be minimised). The considered problem is an instance of  multi-objective optimisation and, as such, is referred to as the multi-objective bulk commodity production problem (MOBCPP) in this paper. This problem is practical and, being an extension of a classic covering problem, belongs to the NP-hard class \cite{Garey1990}. 	
	
	MOBCPP is a multi-objective problem. In multi-objective optimisation we consider $m$ objective functions $f_i(x), i \in \{0,1,\ldots,m-1\}$. Without loss of generality, we may state that the values of all these functions are to be minimised. In this paper, we only consider problems with solutions encoded by $l$ discrete (binary) variables. Thus, a solution $x$ is a binary vector $x=(x_0,x_1,\ldots,x_{l-1})$. For each $x$, the objective value vector is $f(x) = (f_0(x),f_1(x),\ldots,f_{m-1}(x))$.\par

	A method in multi-objective optimisation is expected to return a Pareto front \cite{MoGomeaGecco,EliteArchive}. A Pareto front is a set of non-dominated solutions. A solution $x^0$ \textit{dominates} a solution $x^1$ if and only if $f_i(x^0) \leq f_i(x^1)$ $\forall{i}\in \{0,1,\ldots,m-1\}$ and $f(x^0)\neq f(x^1)$. A solution that is not dominated by any other solution is a Pareto-optimal solution. A Pareto-optimal set $\mathcal{P}_S$ is a set of all Pareto-optimal solutions. The Pareto-optimal front $\mathcal{P}_F$ is a set of objective value vectors of all Pareto-optimal solutions. The number of Pareto-optimal solutions may be large for many problems (in continuous optimisation it is often infinite). Therefore, usually, it is sufficient to find a good approximation of $\mathcal{P}_F$.\par

	If the scale of problem instances is large, metaheuristics may be employed as effective and efficient solvers \cite{paintsOrig,muppetsBaldwinEon,productionYorkRealNumbers,ltgaPopulationSizing}. Evolutionary methods are capable of supporting high-quality solutions consuming a reasonable amount of computation resources. In some practical problems, there are more than one contradicting objectives to optimise. For such problems, instead of a single solution, a set of so-called Pareto optimal solutions is sought. For each of these solutions, improving a single objective causes worsening at least one other objective.\par

	Many methods have been proposed for multi-objective optimisation, for instance, the well-known NSGA-II \cite{nsga2} that employs mechanisms to bias the evolutionary search towards $\mathcal{P}_F$ and preserves the diversity of the final Pareto front approximation. Another proposition is the Multi-Objective Evolutionary Algorithm based on Decomposition (MOEA/D) \cite{moead,Zhou2011,Ma2014b,Ma2014,Ma2016}. The idea behind MOEA/D is to divide a Pareto front and exchange information only between individuals that optimise a similar part of $\mathcal{P}_F$. One of the key advantages of MOEA/D when compared to NSGA-II is that it does not require the computation of so-called crowding distance that is computationally expensive. NSGA-II and MOEA/D are typical reference methods in multi-objective optimisation \cite{MoGomeaSwarm}, so they are also employed as the baseline in this paper.\par

	Solutions for the considered MOBCPP problem are binary-coded. Thus, solution space is a discrete one. The methods that employ linkage learning are particularly effective in the optimisation of problems characterised by such solution spaces. This observation applies to both: theoretical \cite{muppets,3lo,MoGomeaSwarm,ltga} and practical problems \cite{ltgaPopulationSizing,subpopInitLL,mupMemo}. It is also shown that methods employing linkage learning may significantly outperform the other that do not use such techniques \cite{muppets,mohBOA,MoGomeaSwarm,mupMemo}. One of the recent propositions dedicated to solving multi-objective problems is the Multi-objective Gene-pool Optimal Mixing Evolutionary Algorithm (MO-GOMEA) \cite{MoGomeaGecco,MoGomeaSwarm}. MO-GOMEA is based on the concept of the Linkage Tree Genetic Algorithm (LTGA) \cite{ltga,ltgaGomeaNaming,ltgaPopulationSizing}. LTGA is a Genetic Algorithm (GA) that employs linkage learning techniques \cite{3lo,P3Original,ltga,dsmga2} to improve its effectiveness. Similarly to LTGA in the single-objective domains, MO-GOMEA has significantly outperformed competing methods (including NSGA-II and MOEA/D) in multi-objective optimisation \cite{MoGomeaGecco,MoGomeaSwarm}. Among all, to obtain high-quality results, MO-GOMEA clusters the population and processes the subpopulations separately. Therefore, it is capable of optimising different Pareto front parts separately, with the use of linkage that is supposed to describe the features of each Pareto front part. \par
		
	According to \cite{linkageQuality}, the methods that employ linkage learning are dependent on the quality of the linkage they use. If the quality of linkage is too low, linkage-based methods perform similarly to their competitors that do not consider gene-dependencies. In \cite{linkLearningDetermined}, the authors check the dependency between the method's effectiveness and the linkage learning model. They show that if the problem structure is complex (there are many gene dependencies) \cite{watsonHiff,watsonHiffPPSNfirst}, it is favourable to learn linkage during the method execution rather than obtaining the linkage in the pre-optimisation step. Finally, in \cite{3lo}, the authors show that to assure the method's effectiveness, the linkage should be of high quality, but it also should be diverse. To obtain this, they propose a method that utilises a multi-population approach. Note that the multi-population approaches are usually employed to increase population diversity \cite{subpopInitLL,muppets,mupMemo}, but they may also be useful in obtaining a diverse linkage \cite{3lo}. Note that the lack of linkage learning diversity may be a likely reason for a poor performance of LTGA shown in \cite{linkageLearningIsBad}. The research considering the influence of linkage quality on the methods' performance is in its early stage. For instance, it requires further investigation of the reasons why linkage diversity is important to effectively solve problems with complex structure (including so-called overlapping building blocks, which is a typical feature of practical problems) \cite{3lo}. Nevertheless, the objective of this paper is to use the conclusions of the research that has been already made in this area, and to apply these conclusions as intuitions that shall guide us to proposing an effective method for solving the MOBCPP problem. If the intuitions are precise, such a method shall also be effective in solving typical benchmarks employed in a multi-objective optimisation.\par
	
	As stated before, MO-GOMEA is a state-of-the-art method for multi-objective optimisation. However, despite its high effectiveness, it also has some disadvantages. First, it requires a clusterisation of the population. The number of required clusters is adjusted automatically at runtime. However, if the number of clusters is too low or too high, the method may become ineffective \cite{MoGomeaGecco}. Second, although LTGA (the single-objective base of MO-GOMEA) is highly effective in single-objective optimisation for problems with so-called overlapping blocks, it is outperformed by Parameter-less Population Pyramid (P3) \cite{3lo,P3Original,fP3,afP3}. P3 is another state-of-the-art method in single-objective optimisation. The problems with overlapping blocks contain blocks of highly-dependent genes. However, some of the genes in these blocks are also dependent on the genes from other blocks \cite{3lo,P3Original,fP3,afP3}. The feature of inter-block dependencies is typical for practical problems \cite{watsonHiff,watsonHiffPPSNfirst}.\par

	In this paper, we propose a Multi-objective Parameter-less Population Pyramid (MO-P3) to solve the MOBCPP problem effectively. The motivations behind proposing this method are as follows. In contrast to LTGA, P3 maintains numerous different linkages at the same time that should be beneficial for practical problems \cite{3lo}. MO-P3 uses this linkage diversity to omit the necessity of population clusterisation. P3 is a relatively recent method proposition that effectively solves single-objective problems with overlapping blocks. For such problems, P3 has been shown to be significantly more effective than LTGA and Dependency Structure Matrix Genetic Algorithm II (DSMGA-II) \cite{dsmga2}. Since practical problems often contain blocks that overlap \cite{watsonHiff,watsonHiffPPSNfirst}, P3 seems to be a good starting point for solving the practical multi-objective problem considered in this paper. At each MO-P3 iteration, a new individual is added to the population and updated with the use of collected linkages and the rest of the population, similarly to P3. However, in MO-P3, each new individual is assigned a weight vector that directs the search towards a chosen part of the Pareto front. Such a feature may be found similar to the MOEA/D behaviour.\par
	
	The extensive experimental work described in this paper shows that for the considered practical problem, MO-P3 yields results of a higher quality than NSGA-II and MOEA/D. MO-P3 also yields slightly better results than MO-GOMEA. However, its main advantage over MO-GOMEA is that MO-P3 obtains high-quality results significantly faster for MOBCPP (considering both fitness evaluations and computation time). We also present the MO-P3 performance on typical benchmarks. Except for one of them, MO-P3 outperforms all competing methods. Therefore, MO-P3 may be found useful in solving multi-objective problems. Thus, the contribution of this paper is threefold. First, we propose a method dedicated to solving a hard and industrially-relevant practical problem. Second, we fill the gap in the field of Evolutionary Computation that is the lack of a P3-based method dedicated to solving multi-objective problems. Finally, we show that MO-P3 is highly competitive when a typical benchmark set is considered, so we can clearly state that our contribution goes beyond the original problem we set out to solve, and that we propose a new and effective method for multi-objective discrete optimisation.\par

	The rest of this paper is organised as follows. In the next section, we present the related work that includes linkage learning, the presentation of the state-of-the-art methods employing linkage, the issue of Pareto front clusterisation and MO-GOMEA. In Section \ref{sec:problemDef}, we define the MOBCPP problem. In the fourth section, we describe the proposed MO-P3 approach in detail. Sections \ref{sec:exp:paints} and \ref{sec:exp:benchmarks} report the results obtained for MOBCPP and the benchmark problems, respectively. The results are discussed in the seventh section. Finally, the last section points the future research directions and concludes this paper.

	\section{Related Work}
	\label{sec:relWork}
	
	In this section, we present the research related to our propositions. Therefore, in the first subsection, we discuss in detail the issue of linkage learning and linkage learning techniques employed by methods considered in this paper. In Section \ref{sec:relWork:dsmMethods}, we show the details of modern evolutionary methods. These methods are single-objective, but one of them is the base of our proposition and another one is the base of the main competing method considered in this paper. In the fifth subsection, we present the latest advances in discrete multi-objective optimisation. Finally, in the last two subsections, we present MOEA/D in more detail and review previous research related to manufacturing scheduling using multi-objective GAs.

	\subsection{Linkage Learning}
	\label{sec:relWork:ll}
	
	Linkage learning is one of the techniques that are used to detect features of a problem to be optimised. Such knowledge is used during runtime to improve its effectiveness and efficiency. In this section, we present the general linkage classifications and more recent techniques employed by state-of-the-art methods in evolutionary computation. In this paper we concentrate on linkage learning techniques dedicated for discrete domains. However, problem decomposition was found useful also in continuous domains \cite{ieeeSurvey}.

	\subsubsection{General Description and Classifications}
	\label{sec:relWork:ll:general}
	
	Linkage is a piece of information that describes possible dependencies between genes. If such knowledge is accurate and used properly, it may significantly increase the effectiveness of an evolutionary method. In recent years, many different techniques were proposed to obtain linkage. These techniques may be classified with regard to their features. For instance, linkage learning techniques may be classified on the base of: how good and bad linkage are distinguished, how linkage is represented and how linkage is stored \cite{llClassification}. If a method uses only a fitness value to differentiate between a good and bad linkage, it employs a \textit{unimetric way}, which is typical for older Genetic Algorithms (GAs), but some relatively modern methods also adopt it \cite{muppets,muppetsActive}. Nevertheless, current state-of-the-art methods (e.g., Parameter-less Population Pyramid (P3) \cite{P3Original}, Dependency Structure Matrix Genetic Algorithm (DSMGA-II) \cite{dsmga2}, Linkage Tree Genetic Algorithm (LTGA) \cite{ltga}) employ a multi-metric approach, which means that they use more measures than pure fitness to find the linkage of high quality. If the linkage is represented by  dedicated structures (e.g., trees, graphs, matrices or other), such representation is called \textit{virtual}. The linkage represented by the position of the gene in a genotype is called \textit{physical} \cite{muppets}. Finally, the linkage may be stored in one central database or it may be distributed in the population (i.e., each individual may carry its own linkage information).\par
	
	More recent linkage classification was proposed in \cite{DSMorig} and was supplemented in \cite{omidvar,muppetsBaldwinEon}. It considers five different ways of linkage generation. The first class uses a perturbation and analyses the subsequent fitness changes  \cite{omidvar}. Another way to generate linkage is to evolve the order of the genes in the chromosome, which is referred to as \textit{interaction adaptation} \cite{muppets}. An evolutionary method may also build probabilistic models like the Estimation of Distribution Algorithms \cite{mohBOA}. Surprisingly, linkage generated randomly, in some situations, may also improve the method's effectiveness \cite{linkageRandom}. Finally, the last class (proposed in \cite{muppetsBaldwinEon}) is a comparison of evolution results. The methods employing this technique compare the individuals that resulted from different evolutionary processes to obtain linkage. Similar classification of linkage techniques that are employed in Cooperative Coevolution may be found in \cite{linkageClassificationCC}.

	\subsubsection{Dependency Structure Matrix}
	\label{sec:relWork:ll:dsm}
	
	The Dependency Structure Matrix (DSM) is a square matrix that stores the dependencies occurring between the components (genes). This structure is derived from information theory~\cite{dsmga2} and is applied in evolutionary methods to describe gene dependencies. The problem size $n$, where $n$ is a number of genes, determines the size of DSM. Each element $d_{i, j} \in R$ of $\mathrm{DSM} = [d_{i, j}]_{n \times n}$ indicates how significantly the $i^{th}$ and $j^{th}$ genes are dependent on each other. Usually, mutual information~\cite{mutualInformation} is used as the dependency measure. It is defined as 
	\begin{equation}
	\label{eq:mutualInformation}
	I(X, Y) = \sum_{x \in X} \sum_{y \in Y} p(x, y) \ln\frac{p(x,y)}{p(x)p(y)} \geq 0,
	\end{equation}
	where $X$ and $Y$ are random variables. The value of mutual information is proportional to the dependency strength between the pair of genes. If $X$ and $Y$ are independent, the value of $I(X,Y)$ is low, because
	\begin{equation}
	p(x,y) = p(x)p(y) \implies \ln{\frac{p(x,y)}{p(x)p(y)}} = \ln{1} = 0.
	\end{equation}
	It is also assumed that $\ln{\frac{p(x,y)}{p(x)p(y)}}$ equals $0$ when $p(x,y)$, $p(x)$ or $p(y)$ is equal to $0$ as well.

	\begin{table}
		\caption{Population of individuals to demonstrate the DSM creation procedure}
		\centering%
		\label{tab:populationDSM}
		\begin{tabular}{ccccc}
			\hline
			\multirow{2}{*} {\textbf{Population}} & \multicolumn{4}{c}{\textbf{Genotype}}  \\
			& $G_1$ & $G_2$ & $G_3$ & $G_4$ \\
			\hline
			$1^{st}$ individual & 0 & 1 & 0 & 1 \\
			$2^{nd}$ individual & 0 & 1 & 0 & 1 \\
			$3^{rd}$ individual & 1 & 1 & 1 & 1 \\
			$4^{th}$ individual & 1 & 1 & 0 & 1 \\
			$5^{th}$ individual & 0 & 0 & 1 & 1 \\
			\hline
		\end{tabular}
	\end{table} 
	
	To demonstrate the process of DSM creation, we use the population of $5$ binary-coded individuals presented in Table~\ref{tab:populationDSM}, where $G_i$ denotes the $i^{th}$ gene. Formula~(\ref{eq:mutualInformation}) represents the mutual information that can be calculated for any pair of random variables. Particularly, it can be used to measure the dependency between any two genes. Thus, a binary-adjusted version of formula~(\ref{eq:mutualInformation}) may be defined as 
	\begin{equation}
	\label{eq:mutualInformationGenes}
	I(G_i, G_j) = \sum_{g_i \in G_i} \sum_{g_j \in G_j} p_{i,j}(g_i, g_j) \ln\frac{p_{i,j}(g_i, g_j)}{p_i(g_i)p_j(g_j)},
	\end{equation}
	where $g_i$ and $g_j$ indicate the possible values of the $i^{th}$ ($G_i$) and $j^{th}$ ($G_j$) genes, respectively. For instance, if an optimisation problem is binary then $g_i \in \{0, 1\} = G_i$ and $g_j \in \{0, 1\} = G_j$. To calculate the probabilities presented in formula~(\ref{eq:mutualInformationGenes}), all individuals in a population are taken into consideration. The $p_{i,j}(g_i, g_j)$ value denotes the joint probability that a value of the $i^{th}$ gene is $g_i$ and the $j^{th}$ gene has value $g_j$ simultaneously. Moreover, $p_i(g_i)$ is used to indicate the probability that a value of the $i^{th}$ gene is $g_i$. Table~\ref{tab:DSM} presents DSM obtained for the population shown in Table~\ref{tab:populationDSM}. All DSM entries presented in Table~\ref{tab:DSM} have been calculated using formula~(\ref{eq:mutualInformationGenes}).
	
	\begin{table}
		\caption{DSM for the population presented in Table~\ref{tab:populationDSM}}
		\centering
		\label{tab:DSM}
		\begin{tabular}{ccccc}
			\hline
			\textbf{} & $G_1$ & $G_2$ & $G_3$ & $G_4$ \\
			\hline
			$G_1$ & X & 0.12 & 0.00 & 0.00 \\
			$G_2$ & 0.12 & X & 0.22 & 0.00 \\
			$G_3$ & 0.00 & 0.22 & X & 0.00 \\
			$G_4$ & 0.00 & 0.00 & 0.00 & X \\
			\hline
		\end{tabular}
	\end{table} 
	
	DSM-based linkage learning may lead to excellent results and is employed by leading methods in the field of discrete optimisation \cite{P3Original,ltga,psDSMGA2,dsmga2}. For some problems, it facilitates finding a high-quality linkage \cite{ltga}, which is crucial to solve the problem. Additionally, it aids updating the linkage information during runtime, which is key when addressing problems with complex structure \cite{linkLearningDetermined}. Such a structure may be commonly found in practical problems \cite{watsonHiff,watsonHiffPPSNfirst}.\par
	
	As presented in \cite{linkageQuality}, not all problem types are easy to decompose for a DSM-based linkage learning. Recently, Linkage Learning based on Local Optimisation (3LO) was proposed in \cite{3lo}. 3LO is an empirical linkage learning technique, which means that the dependencies \textit{predicted} by other linkage learning techniques are replaced based on an empirical check. Thanks to the idea behind it, 3LO is proven not to report any \textit{false linkage}. The \textit{false linkage} takes place when two independent genes are pointed as being dependent by a linkage learning technique. A drawback of 3LO is its computational cost. Therefore, the methods using it perform worse when overlapping problems need to be solved \cite{3lo}. This observation justifies the choice of a DSM-using method as the base of our proposition.

	\subsubsection{Linkage Trees}
	\label{sec:relWork:ll:linkTree}
	
	DSM has been created to find linkage and it contains only pairwise gene-dependency values. Therefore, a clustering algorithm is employed to merge pairs of genes into larger groups. Different techniques of DSM utilisation were proposed~\cite{dsmga2, dsmga2e, ltga, P3Original}. The only technique employed by methods considered in this paper is the linkage tree construction algorithm and hence it is described in details below. Nevertheless, at the end of this section, we also give some insights into other DSM utilisation techniques.\par
	
	To construct a linkage tree, the distance $D(G_i, G_j)$ between the $i^{th}$ and $j^{th}$ genes is calculated using mutual information (formula~(\ref{eq:mutualInformationGenes})) and joint entropy:
	
	\begin{equation}
	\label{eq:distanceMeause}
	D(G_i, G_j) = \frac{H(G_i, G_j) - I(G_i, G_j)}{H(G_i, G_j)},
	\end{equation}
	where
	\begin{equation}
	\label{eq:entropy}
	H(G_i, G_j) = - \sum_{g_i \in G_i} \sum_{g_j \in G_j} p_{i,j}(g_i, g_j) \ln{p_{i,j}(g_i, g_j)}.
	\end{equation}
	
	Note that $H(G_i, G_j)$ equal $0$ implies that distance $D(G_i, G_j)$ is $0$ as well. In Table~\ref{tab:distances}, we report values of gene distances computed for the population presented in Table~\ref{tab:populationDSM}.
	
	\begin{table}
		\caption{Distances between genes for the population presented in Table~\ref{tab:populationDSM}}
		\centering
		\label{tab:distances}
		\begin{tabular}{ccccc}
			\hline
			\textbf{} & $G_1$ & $G_2$ & $G_3$ & $G_4$ \\
			\hline
			$G_1$ & X & 0.88 & 1.00 & 1.00 \\
			$G_2$ & 0.88 & X & 0.76 & 1.00 \\
			$G_3$ & 1.00 & 0.76 & X & 1.00 \\
			$G_4$ & 1.00 & 1.00 & 1.00 & X \\
			\hline
		\end{tabular}
	\end{table} 
	
\begin{figure}
		\centering
		\includegraphics[width=0.75\linewidth]{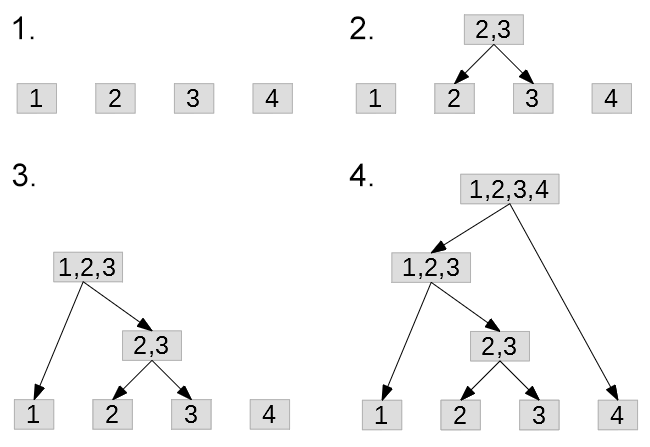}
		\caption{Subsequent steps of the linkage tree creation process for the population from Table~\ref{tab:populationDSM}}
		\label{fig:linkageTreeCreation}
	\end{figure}

	A linkage tree consists of the nodes corresponding to the clusters which group the genes that are considered to be dependent on one another. During linkage tree construction, the two most related clusters are joined. Initially, the clusters containing one consecutive single gene are created. Thus, the linkage tree construction algorithm creates $n$ single-gene clusters, where $n$ is the given optimisation problem size. Then, the merging operation is repeated until only one cluster (consisting of all genes) remains. Formula~(\ref{eq:distanceMeause}) is used to calculate the distance between two clusters which contain only a single gene. If one of the clusters contains more than one gene, the following reduction formula is used:
	\begin{equation}
	D(C_k, (C_i \cup C_j)) = \frac{|C_i|}{|C_i| + |C_j|} D(C_k, C_i) + \frac{|C_j|}{|C_i| + |C_j|} D(C_k, C_j),
	\end{equation}
	where $|C_i|$, $|C_j|$ and $|C_k|$ indicate the sizes of clusters $C_i$, $C_j$ and $C_k$, respectively. According to Table~\ref{tab:distances}, the distance between clusters $\{G_1\}$ and $\{G_2, G_3\}$ is calculated as follows:
	\begin{equation}
	\begin{aligned}
	D(\{G_1\}, (\{G_2\} \cup \{G_3\}))  &= \frac{|\{G_1\}|}{|\{G_2\}| + |\{G_3\}|} D(\{G_1\}, \{G_2\}) 
	\\ &+ \frac{|\{G_3\}|}{|\{G_2\}| + |\{G_3\}|} D(\{G_1\}, \{G_3\})
	\\ &= \frac{0.88}{2} + \frac{1}{2} = 0.94.
	\end{aligned}
	\end{equation}
	The process of the linkage tree creation for DSM given in Table~\ref{tab:populationDSM} is presented step by step in Figure~\ref{fig:linkageTreeCreation}. To simplify the diagram, indication $G_i$ has been replaced by number $i$. For instance, in Figure~\ref{fig:linkageTreeCreation}, we use $1$ instead of $G_1$.
	
	Linkage trees are employed by Linkage Tree Genetic Algorithm (LTGA) \cite{ltga}, also denoted as Linkage Tree Gene-pool Optimal Mixing Evolutionary Algorithm (LT-GOMEA) \cite{ltgaGomeaNaming}. Another method that employs linkage trees is Parameter-less Population Pyramid (P3) \cite{P3Original,fP3}. Both methods are described in the next subsection.\par

	Another way of using DSM is creation of an incremental linkage set. The incremental linkage set consists of sequences of gene starting indexes. During the process of gene-sequence creation, a single gene index is selected randomly. Then, the index that has the strongest relation to the last gene in the sequence and is not included in the sequence is added to the sequence. Incremental linkage sets are employed by Dependency Structure Matrix Genetic Algorithm II (DSMGA-II) \cite{dsmga2} and Two-edge Dependency Structure Matrix Genetic Algorithm II (DMSGA-IIe) \cite{dsmga2e} that are presented in the next section.

	\subsection{DSM-using Methods}
	\label{sec:relWork:dsmMethods}
	
	In this section, we present different methods that employ DSM and information theory for linkage discovery.

	\subsubsection{Linkage Tree Genetic Algorithm}
	\label{sec:relWork:dsmMethods:ltga}
	
	Linkage trees have been employed by Linkage Tree Genetic Algorithm (LTGA)~\cite{ltga, ltgaPopulationSizing}, one of the first methods using DSM which has been shown to be highly effective. LTGA is a population-based method that uses the linkage tree construction algorithm described in Section \ref{sec:relWork:ll:linkTree}. Recently, LTGA has been improved and renamed to Linkage Tree Gene-pool Optimal Mixing Evolutionary Algorithm (LT-GOMEA) \cite{ltgaPopulationSizing}. \par
	
	Instead of crossover, LTGA uses the operator called optimal mixing (OM). During OM, two individuals (called \textit{source} and \textit{donor}) and a cluster (a node from a linkage tree) are involved. The genes from the donor individual that are marked by the cluster replace the appropriate genes in the source individual. The operation is reversed if the fitness of the source decreases. Otherwise, the source remains modified. All individuals in the population are mixed using OM. During OM, all clusters except the linkage tree root are considered. The donor is selected randomly for each cluster. If, after OM, an individual remains unmodified, OM is executed for this individual once again with the best-found individual as the donor. This step is called the force improvements (FI) phase. The second situation in which FI is executed takes place when the best-found individual has not been improved for a certain number of iterations. \par
	
	As an example of OM, let us consider a 6-bit binary problem. The genotype of a source individual is $110011$, and its fitness is 6. The first considered cluster marks genes 1,2, and 5, and the donor individual is $010101$. After mixing, the genotype of the donor individual will be $010001$, and its fitness will decrease to 4. Therefore, the change introduced by mixing is rejected (we wish to maximise the fitness). The second considered cluster marks genes 2 and 3, and the randomly chosen donor individual for this cluster is $000111$. After mixing, the donor's genotype will be $000101$, and its fitness will be 6. Since fitness has not decreased, the change is preserved. The third cluster marks genes from 2 to 5, and the individual chosen for this cluster is $111111$. After mixing, the genotype of the donor will be $011111$, and its fitness will be 7. Therefore, the change will be preserved. The operation of OM will continue in the manner shown above until all the clusters that do not cover the whole genotype will be considered.\par
	
	LTGA requires one parameter, namely the population size. Finding its appropriate value for a particular test case via tuning may be difficult. Therefore, a population-sizing scheme for LTGA was proposed in~\cite{ltgaPopulationSizing}. LTGA employing this scheme is denoted as LT-GOMEA. LT-GOMEA maintains multiple LTGA instances with different population sizes. The first LTGA instance contains only one individual. During LT-GOMEA execution, new LTGA instances with a doubled population size are added at every $4^{th}$ iteration. Some of the LTGA instances may be found useless and deleted, which limits their number. A single LTGA instance is found useless if all of its individuals are the same or its average population fitness is worse than the average fitness of at least one LTGA with a larger population size. Additionally, all LTGA instances with a smaller population than the LTGA instance found useless are treated as useless as well. All LTGA instances are isolated from each other. Only during the FI phase, the globally best individual (found by any LTGA instance) is used as a donor. Additionally, LT-GOMEA introduces two changes to LTGA. First, LT-GOMEA computes DSM on the base of the whole population, while LTGA uses only a half of the population. Second, during the OM operation, LT-GOMEA considers linkage tree clusters in a random order, while LTGA uses them in the order of their creation. These two changes are supposed to increase the quality of linkage and remove the potential bias that may influence the method for a particular problem, respectively.\par

	\subsubsection{Parameter-less Population Pyramid}
	\label{sec:relWork:dsmMethods:p3}

	Parameter-less Population Pyramid (P3)~\cite{P3Original} uses the same linkage tree construction algorithm and the OM operator as LTGA. However, the population structure is significantly different from any other GA-based method. P3 maintains its population in a pyramid-like structure divided into subpopulations called \textit{levels}. Every individual in the population is unique. The population size is not limited and increases during runtime.\par
	
\begin{figure}
		\centering
		\includegraphics[width=0.9\linewidth]{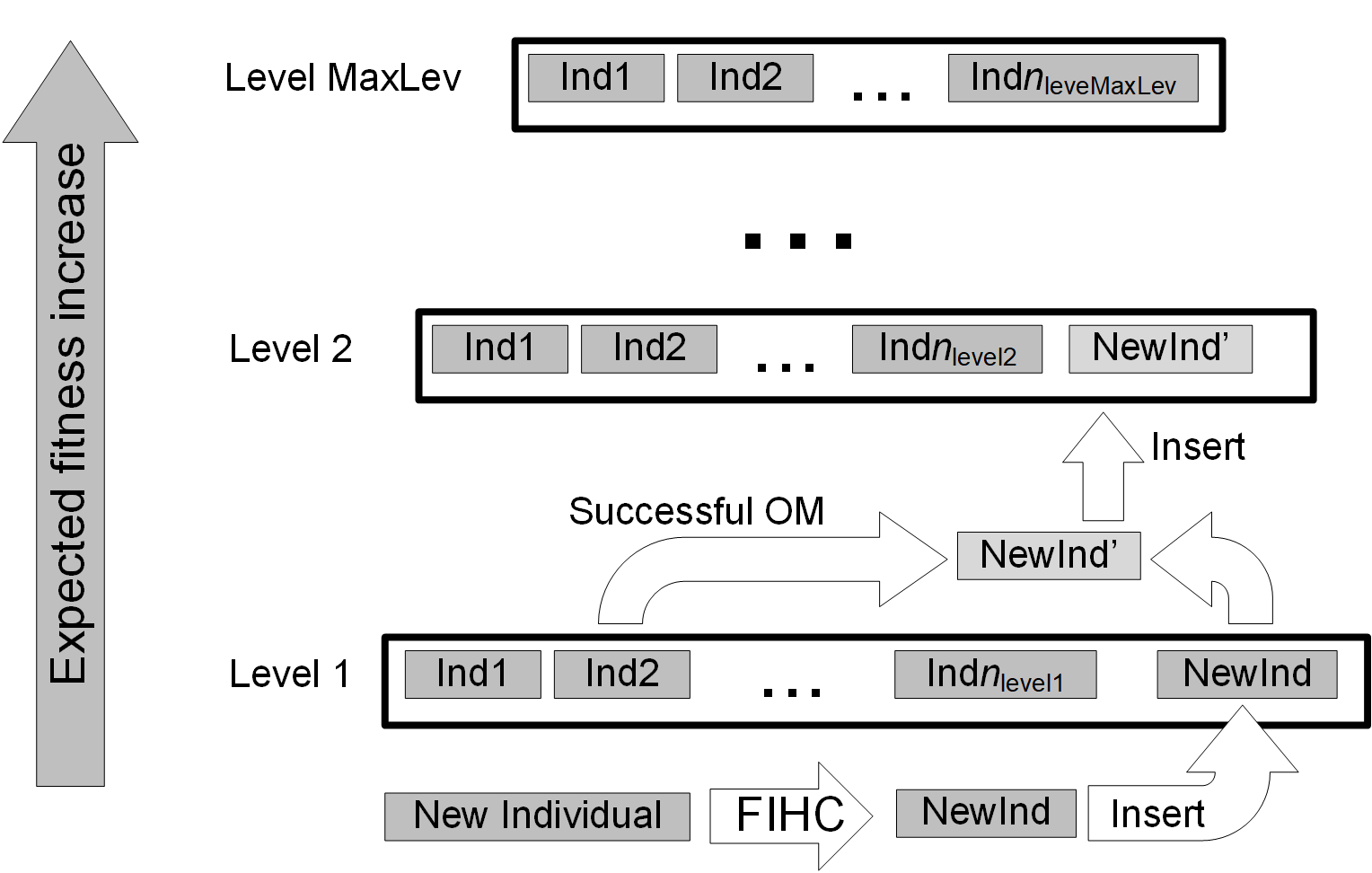}
		\caption{P3 idea visualization}
		\label{fig:p3FlowChart}
	\end{figure}

	The general P3 procedure can be described in the following way. At every iteration, a new individual is created randomly and initially optimised by First Improvement Hill Climber (FIHC)~\cite{P3Original}. FIHC is a local search algorithm operating on vector $\overrightarrow{x}$ of $n$ decision variables, $\overrightarrow{x}=[x_1, \ldots , x_n]$. Initially, FIHC randomly chooses a gene order. For each gene $x_i$, all available values are checked until a fitness improvement is found. If so, then the original $x_i$ value is replaced. This procedure is executed until no gene is changed during a FIHC iteration. After the optimisation is done by FIHC, the new individual climbs up the pyramid. With the use of OM, it is mixed with all individuals of a single level. The bottom pyramid levels are considered first. If a fitness of the new individual is improved during this operation, a new individual is added to the pyramid level. If a successful OM involves an individual from the top-level, a new \textit{level} (subpopulation) is added to the pyramid. The overall idea of P3 work is presented in Figure \ref{fig:p3FlowChart}.\par

	\subsubsection{Dependency Structure Matrix Genetic Algorithm II}
	\label{sec:relWork:dsmMethods:dsmga2}
	
	Dependency Structure Matrix Genetic Algorithm II is another method that employs DSM-based linkage learning \cite{dsmga2}. However, unlike P3 or LTGA, it uses an incremental linkage set (ILS) instead of a linkage tree. ILS is a sequence of gene indexes. The process of ILS building starts from a single gene index and adds a new one with the strongest connection to the previously added gene (in terms of the DSM weights) that has not been included in the sequence yet. Similarly to LTGA, DSMGA-II maintains a single population with a fixed number of individuals. Two operators are used: restricted mixing and back mixing. Restricted mixing is used to process a single individual. First, an incremental linkage set is created, starting from a random gene index. Then, the consecutive genes are flipped according to the incremental linkage set. The operation is maintained until a better fitness is obtained or until the same fitness is obtained, but the modified individual is absent in the population. If all the genes have been flipped but the fitness of the modified individuals is worse than that of the starting individual, the changes done by restricted mixing are rejected. However, if restricted mixing leads to a change (i.e., the new individual has a higher or the same fitness but with a genotype that does not exist in the population), the back mixing operation is triggered. During back mixing, the change introduced by restricted mixing is injected into other individuals. The injection is preserved if it improves fitness and rejected otherwise.\par
	
	DSMGA-II has been shown to be effective when solving theoretical \cite{dsmga2} and practical problems \cite{steinerTrees}. Recently, its parameter-less version that employs a population-sizing scheme (denoted as psDSMGA-II) has been proposed in \cite{psDSMGA2}. Although psDSMGA-II improves the effectiveness of the original DSMGA-II, its main disadvantage is the same as for its predecessor: it is less effective in solving the problems with overlapping building blocks in comparison with LT-GOMEA or P3 \cite{afP3,psDSMGA2}.

	\subsection{Linkage Diversity}
	\label{sec:relWork:linkageDiversity}
	
	P3 is effective for solving hard computational problems \cite{P3Original,fP3,afP3}. Compared to LT-GOMEA, P3 performs significantly better when the problem to be solved has highly overlapping blocks (e.g., NK fitness landscapes) \cite{afP3}. This feature is important in practice because it is typical for real-life problems \cite{watsonHiffPPSNfirst,OverlappingSimon}. To the best of our knowledge, no detailed analysis of P3 superiority over LT-GOMEA and DSMGA-II in solving problems that overlap has  been performed and published yet. Below, we propose an explanation of this superiority.\par
	
	Let us introduce the deceptive function of unitation~\cite{decFunc}. Formula (\ref{eq:dec3}) defines the deceptive function of order $k$, the solution is binary-coded (i.e., is a string of $0$s and $1$s).
	
	\begin{equation}
	\label{eq:dec3}
	\mathit{dec(u)}=
	\begin{cases}
	k - 1 - u & \text{if } u < k\\
	k & \text{if } u = k
	\end{cases},
	\end{equation}
	where $u$ is a sum of gene values (so called \textit{unitation}) and $k$ is the deceptive function size.
	
\begin{figure}
		\centering
		\includegraphics[width=0.5\linewidth]{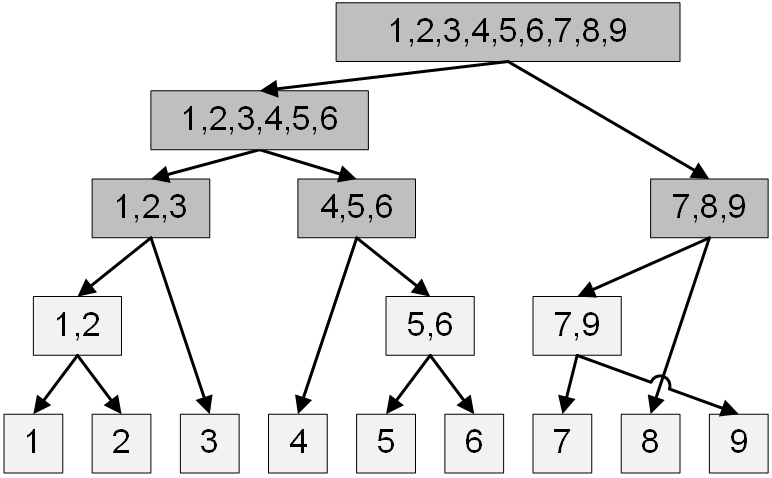}
		\caption{Linkage tree that represents a perfect linkage for the concatenation of three order-3 deceptive functions}
		\label{fig:perfectLinkageTree}
	\end{figure}

	The optimal solution of the order-3 deceptive function is $111$, while the suboptimum is $000$. Let us consider the concatenation of three order-3 deceptive functions, where the first three bits refer to the first function (building block), the second three bits refer to the second function and so on. The optimal solution to this problem is $111 111 111$. However, most of the population of typical GA-based methods is deceived to the $000 000 000$ solution. If this problem is sufficiently large, it may become intractable. On the other hand, deceptive functions' concatenations are easy to be solved if the problem nature (the perfect linkage) is known \cite{grayWhitley}. In the given example, the perfect linkage that groups the dependent gene indexes is $(1,2,3)$, $(4,5,6)$ and $(7,8,9)$. Such linkage may be represented by a single linkage tree (Figure \ref{fig:perfectLinkageTree}).
	
	\begin{figure}
		\centering
		\includegraphics[width=0.3\linewidth]{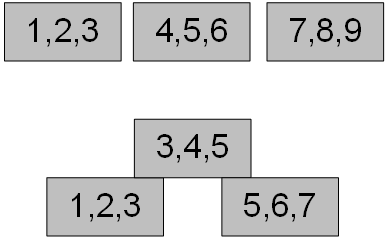}
		\caption{Block positions dependencies for the problem constructed from the concatenation of three order-3 deceptive functions without overlap and with overlap $o=1$}
		\label{fig:posDependencies}
	\end{figure}
	
	Let us now consider the problem of overlapping deceptive functions that is an example of a problem with overlapping blocks. The size of overlap is defined by $o\in\{0,1,\ldots,k-1\}$, where $k$ is a length of all the considered blocks. The first block is defined on the first $k$ positions of the genotype. All blocks except the first one are defined on the last $o$ positions of the preceding block and the next $k-o$ positions. For instance, the positions referring to the second block start at the $(k-o+1)^{th}$ position and finish at the $(2\cdot k-o)^{th}$ position. The examples of deceptive blocks concatenations with and without overlap are given in Figure \ref{fig:posDependencies}.
	
	\begin{figure}[h]
		\subfloat[Perfect linkage for the first block on positions $\{1,2,3\}$]{%
			\includegraphics[width=0.3\linewidth]{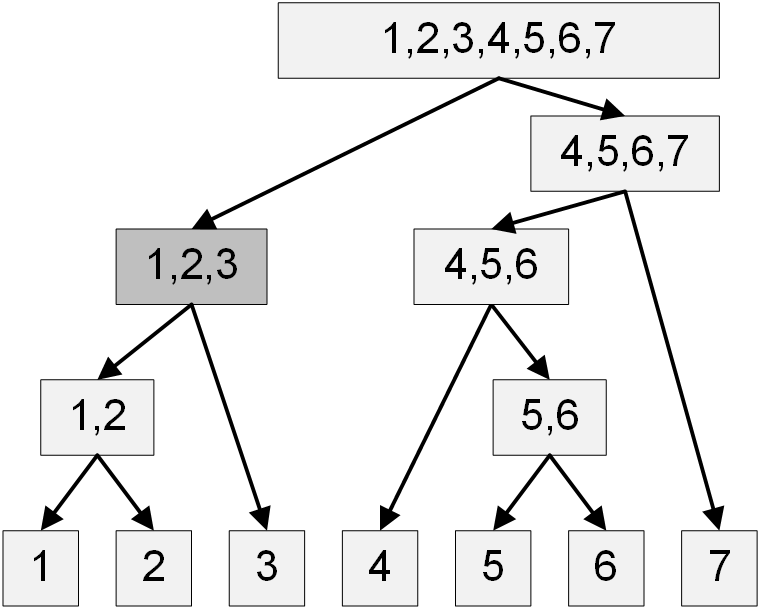}}
		\label{fig:linkageForOverlap_a}\hfill
		\subfloat[Perfect linkage for the last block on positions $\{3,4,5\}$]{%
			\includegraphics[width=0.3\linewidth]{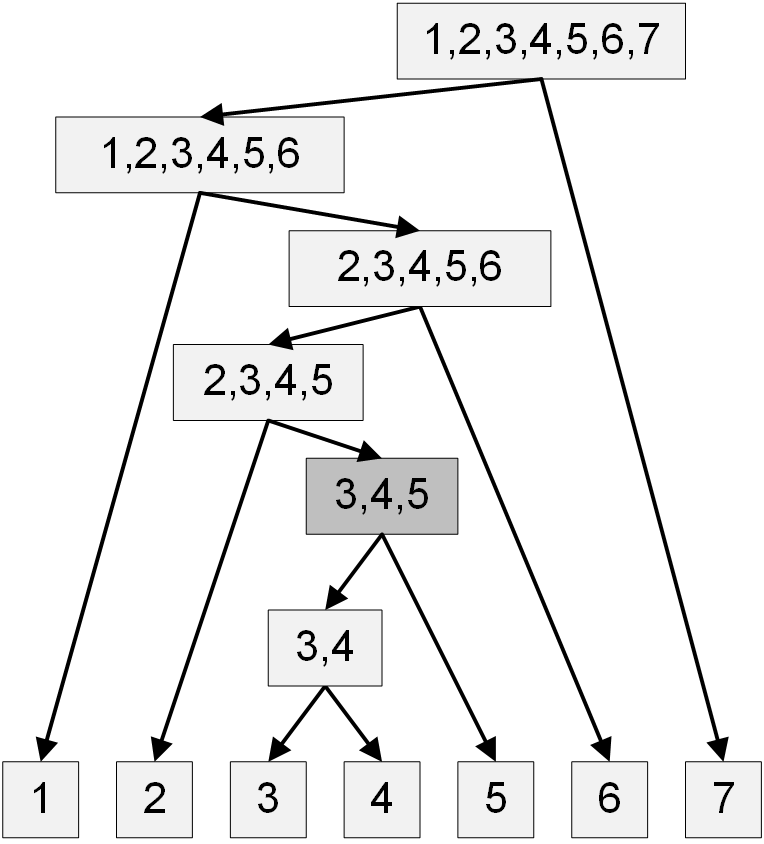}}
		\label{fig:linkageForOverlap_b}\hfill
		\subfloat[Perfect linkage for the last block on positions $\{5,6,7\}$]{%
			\includegraphics[width=0.3\linewidth]{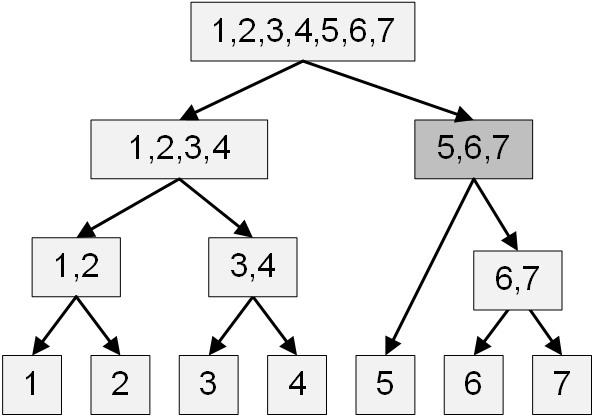}}
		\label{fig:linkageForOverlap_c}\hfill
		\caption{Possible linkage trees for three order-3 deceptive functions concatenation with overlap $o=1$}
		\label{fig:linkageForOverlap}
	\end{figure}
	
	In Figure \ref{fig:linkageForOverlap}, we present possible linkage trees for the concatenation of three order-3 deceptive functions with the $o=1$ overlap. Note that although all the linkage trees are correct, each of them marks only one of the blocks. If the blocks overlap, it is impossible to mark all blocks with a single tree. Therefore, maintaining and using several different linkages at the same time may be beneficial when solving problems with overlaps This observation is confirmed by the results presented in \cite{3lo}. The proposed reasoning is valid only under the assumption that a single linkage tree (even if it is correct) may not be enough to solve the problem with overlaps. To the best of our knowledge, the study that analyses the need for linkage diversity has not been proposed yet, but the results presented in the literature seem to support the above claim \cite{3lo,P3Original,fP3,afP3,psDSMGA2}. The comparison between the original DSMGA-II and DSMGA-II with population-sizing (psDSMGA) show that psDSMGA-II significantly outperforms its predecessor for overlapping problems \cite{dsmga2PopulationSizing}. A similar observation can be made for LTGA and LT-GOMEA comparison \cite{P3Original,afP3}. Population-sizing was proposed to eliminate the necessity of tuning and defining the population size parameter (see Section \ref{sec:relWork:dsmMethods:ltga}). However, as a side effect, population-sizing leads to the maintenance of more than one LTGA/DSMGA-II population. All these populations maintain separate linkages and communicate with each other via the global-best individual. Thus, the globally best individual may be updated by OM in which the donor may be any individual from any LTGA/DSMGA-II population. Depending on the population, a different linkage is used. The above reasoning leads to the conclusion that population-sizing, as a side-effect, introduces a linkage diversity (limited but it is still better than none) and this linkage diversity seems to lead to the results' quality improvement for problems with overlapping building blocks.\par
	
	\begin{figure}
		\centering
		\includegraphics[width=0.95\linewidth]{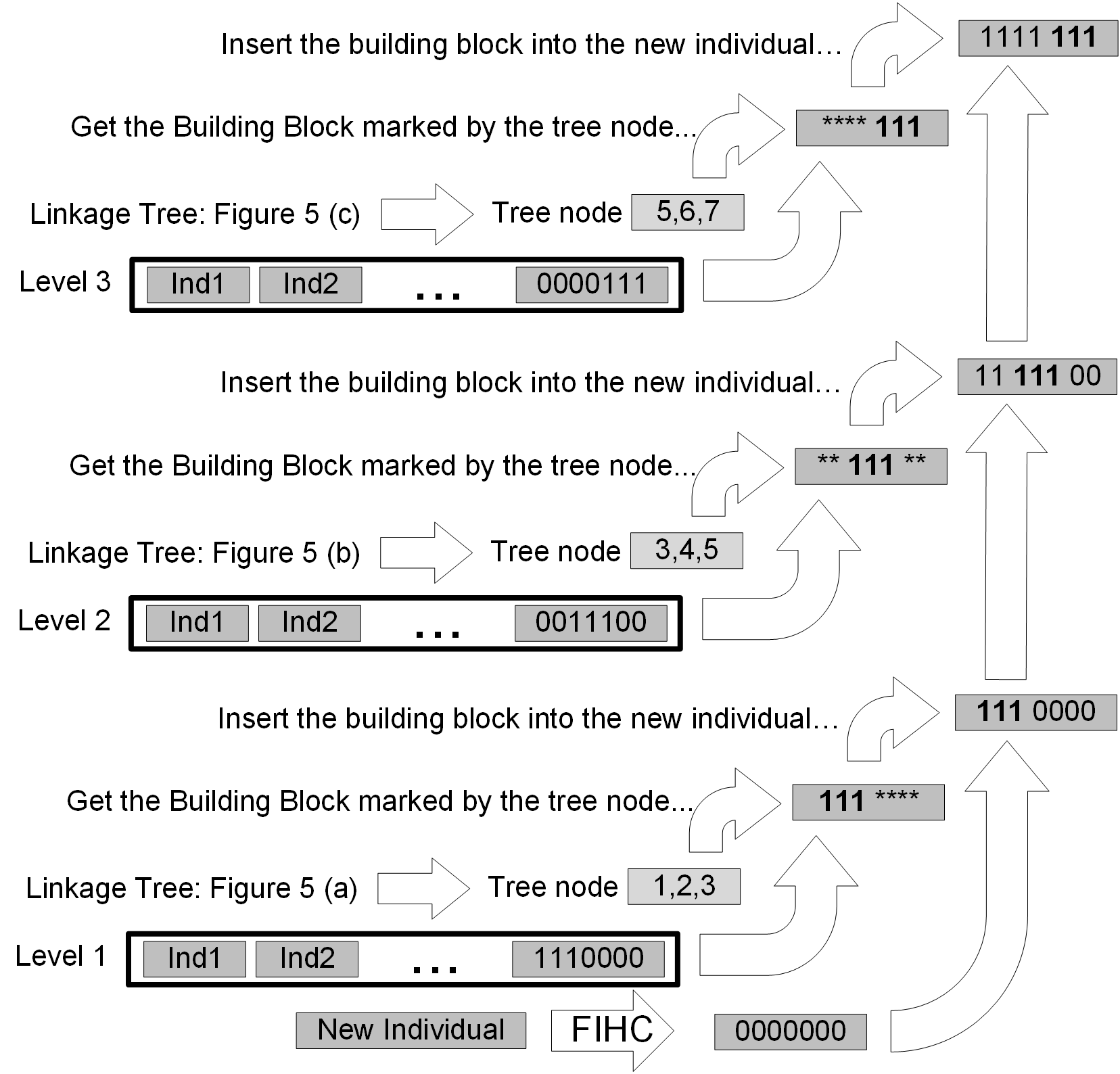}
		\caption{The example of linkage diversity employment in P3}
		\label{fig:linkDiversP3Example}
	\end{figure}

\subsection{The significance of Linkage Quality and Diversity}
\label{sec:relWork:linkageDiversityExample}
	
As presented in \cite{linkageQuality}, the quality of the linkage may be the key to solve hard computational problems. To show the significance of using a diverse linkage, let us analyze an example shown in Figure \ref{fig:linkDiversP3Example}. We consider a problem, built from three order-3 deceptive functions with overlap $o=1$, presented in the lower part of Figure \ref{fig:posDependencies}. The P3-like population is divided into three levels. The linkage information of the first, second, and third level corresponds to the Linkage Trees presented in Figure \ref{fig:linkageForOverlap} (a), (b), and (c), respectively. We assume that, among all, the pyramid contains the following individuals:

\begin{itemize}
	\item Individual $1110000$ (optimal for the first block), on the first level
	\item Individual $0011100$ (optimal for the second block), on the second level
	\item Individual $0000111$ (optimal for the third block), on the third level
\end{itemize}
	
In the example pictured in Figure \ref{fig:linkDiversP3Example}, we analyze a single iteration of P3, in which we try to add a new individual to the pyramid with the genotype $0000000$. It is possible that after OM with the individuals on the first level, the new individual will receive the optimal value for the first block (after that, the genotype of the new individual will be $1110000$). If during the OM with the second and third level, the new individual will receive the optimal value for the second and the third block, respectively, then the final genotype of the new individual will be optimal (built only from $1$s). Note that it is possible to obtain an optimal individual because P3 employs many different linkages that mark various parts of the genotype.\par

Let us now consider the same population of individuals but grouped in a single population (such population contains individuals $1110000$, $0011100$, and $0000111$). We assume that the linkage corresponds to the Linkage Tree presented in Figure \ref{fig:linkageForOverlap} (b). Note that in such a situation, it is impossible to obtain the optimal individual using OM. It is possible to insert the second block of $1$s from individual $0011100$ to individuals $1110000$, $0000111$, and obtain individuals $1111100$, $0011111$, respectively. However, using the Linkage Tree from Figure \ref{fig:linkageForOverlap} (b), it is impossible to pass the first and the third block of $1$s successfully, without destroying the other blocks. Note that it is impossible to separately insert $1$s for genes 1, 2, 6, and 7 – the fitness value of $1011111$, $0111111$, $1111101$, and $0111110$ is $6$ and is lower than the fitness value of $0011111$ and $1111100$ that is 7. Thus, an operation converting a single $0$ into $1$, for individuals $0011111$ and $1111100$, will be rejected by OM.\par

For problems encoded with a large number of genes and with a high amount of overlaps, the dependencies between genes will be significantly more complex. Thus, it is intuitive that, in such cases, it is preferable to use more diverse linkage information. LT-GOMEA uses a population-sizing mechanism. Thus, it also maintains many subpopulations, and therefore, it also maintains many linkages. However, based on the research results presented in \cite{linkageQuality}, the number of levels in the pyramid is usually significantly higher than the number of subpopulations maintained by LT-GOMEA. Thus, when P3 and LT-GOMEA are applied to solve the problem, we may expect that P3 will use a more diverse linkage than LT-GOMEA. Therefore, P3 is more suitable to solve overlapping problems.\par

In this paper, as a competing method, we also consider NSGA-II that employs a standard crossover operator and no linkage learning mechanisms. Let us consider the probability to successfully insert the first block of three $1$s from individual $1110000$ to individuals $0011100$ and $0011111$. We consider the uniform crossover that is independent of the gene order. The assumption that genetic operators should be independent of gene order seems intuitive and is typical for modern evolutionary algorithms \cite{3lo}. The probability of successful insertion of the first block of $1$s to individual $0011100$ is $2^{-4}$ (we need to exchange genes 1 and 2, and we shall not exchange genes 4 and 5). However, if we wish to insert the same block of $1$s to individual $0011111$, the probability will be $2^{-6}$. Moreover, the probability to successfully exchange a particular building block without destroying the other blocks will decrease quickly with the increase of the problem size. Therefore, the typical crossover operators that do not use linkage learning will not be effective when strong inter-gene dependencies exist.

As shown above, the diverse linkage maintained by P3 may be highly useful when overlapping problems are being solved. However, the pyramid-like form of the P3 population has some drawbacks. For instance, when the number of levels is large (eg., over 30), it may be hard to exchange some building blocks, even if appropriate linkage and appropriate blocks exist in the population. This issue has been detected for non-overlapping problems, and the modifications of P3 to address this issue were proposed in \cite{fP3,afP3}.

	\subsection{Pareto Front Clusterisation}
	\label{sec:relWork:clusterisation}
	
	The goal of multi-objective optimisation is to obtain a Pareto front that covers or is relatively close to the optimal Pareto front. Thus, to measure the quality of a Pareto front, its proximity and diversity are often compared with the optimal Pareto front \cite{frontQualityMeasurement}. To obtain a diverse and high-quality Pareto front, evolutionary methods tend to preserve the overall population diversity, often emphasising the diversity in the objective space. This may be achieved by employing the crowding distance like in NSGA-II \cite{nsga2} or a density measure like in SPEA2 \cite{spea2}. Nevertheless, such operators may not be sufficient to obtain good quality results for hard optimisation problems. Therefore, linkage learning techniques may be useful to recognise the problem's nature and exchange appropriate solution parts during the optimisation process \cite{MoGomeaGecco,MoGomeaSwarm}. However, for different Pareto front parts, the problem's features may be significantly different. For instance, for the practical problem considered in this paper, the solution minimising the production makespan may be significantly different from the solution minimising production surpluses. Similar observations can be made for other practical problems \cite{MoGomeaSwarm}. Therefore, some of the evolutionary methods that are employed for the multi-objective optimisation split the population into clusters, usually based on the objective space \cite{mohBOA,clusterizationBosman}.

	\subsection{Multi-objective Gene-pool Optimal Mixing Evolutionary Algorithm}
	\label{sec:relWork:moGomea}
	
	MO-GOMEA is a multi-objective version of LTGA (see Section \ref{sec:relWork:dsmMethods:ltga}). Except for the concept of LTGA, it employs other ideas, like Pareto front clusterisation and elitist archive. MO-GOMEA is a parameter-less method, which is an important feature for practical purposes.\par
	
	MO-GOMEA uses a so-called \textit{elitist archive}. The elitist archive is a separate population where non-dominated solutions are stored. Maintaining such a buffer is beneficial for multi-objective evolutionary algorithms because, during the search, some non-dominated solutions may be discarded due to the stochastic nature of the search \cite{MoGomeaSwarm}. In some optimisation problems, the Pareto front may contain an infinite (or too large to store) number of solutions \cite{ElitistArchive2}. Thus, if a new non-dominated solution is found, it shall be added to the elitist archive if it dominates at least one solution from the elitist archive or if it increases the diversity of the archive in the solution space \cite{ElitistArchive2}.\par
	
	At each method iteration, MO-GOMEA clusters its population with the use of \textit{k}-leader-means clustering \cite{clusterizationBosman}. The population is divided into \textit{k} clusters containing \textit{c} solutions. In MO-GOMEA $c = \frac{2}{k}\cdot|\mathcal{P}_F|$; such a size of \textit{c} causes the clusters to overlap and avoid the situation in which some individuals do not belong to any cluster. Each cluster is a subpopulation that is later processed by LTGA. The only difference is that since the problem is multi-objective, the source solution is found improved if, after the optimal mixing, the altered source solution dominates its previous version or it can be added to the elitist archive. MO-GOMEA requires specifying two parameters: the population size and the number of clusters. To overcome this issue, it employs the so-called interleaved multi-start scheme (IMS) \cite{MoGomeaSwarm} that is equivalent to the population-sizing scheme described in Section \ref{sec:relWork:dsmMethods:ltga}.\par

	\subsection{Multi-Objective Evolutionary Algorithm based on Decomposition}
	\label{sec:relWork:moead}
	
 Multiobjective Evolutionary Algorithm based on Decomposition (MOEA/D) is a multi-objective optimisation algorithm whose main principle is to decompose an $m$-objective problem into $N$ single-objective subproblems \cite{moead}. The three decomposition techniques in MOEA/D are: the weighted sum approach, the Tchebycheff approach, and the penalty-based boundary intersection approach. For example, employing the Tchebycheff approach, maximisation of an $m$-objective optimisation problem $F(x) = (f_1(x), \ldots, f_m(x))^T$ can be decomposed to the optimisation of the $N$ subproblems and the objective function of the $j$-th subproblem, $j=1,\ldots,N$, is:

\begin{equation} \label{tchebycheff}
g^{te} \left(x \vert \lambda^j, z^*\right) = \max_{1 \leq i \leq m} \left\{ \lambda_i^j | f_i(x) - z^*_i | \right\},
\end{equation}
where $x$ indicates a given solution in the decision space, $\lambda^j=(\lambda_1^j,\ldots,\lambda_m^j)$ is a weight vector, $z^*$ indicates a set of reference points $z_i^* = min \{f_i(x) | x \in \mathrm{\Omega} \}$ and $|\cdot|$ denotes the Euclidean distance.  Several different methods for selecting weight vectors $\lambda^j$ have been proposed, including classic simplex-centroid and simplex-lattice, as well as more recent transformation methods and uniform decomposition measurement \cite{Ma2014c}.
	
	Single-objective subproblems (\ref{tchebycheff}) are solved simultaneously, employing evolutionary algorithms. Each solution to such single-objective subproblems forms a Pareto-optimal front of the original multi-objective problem $F(x)$.
	
	One of the main assumptions of MOEA/D is that the optimal solutions to neighbouring subproblems (in terms of the Euclidean distance between the corresponding weight vectors) are likely to be similar. Hence, a chromosome describing a solution to a certain single-objective subproblem can be crossed-over only with the chromosomes of the neighbouring subproblems, where the neighbourhood size is a parameter. Each population comprises the best solutions found so far to all $N$ subproblems. In MOEA/D, the one-point crossover operator and the standard mutation operator are applied. Thanks to defining the weight vector set a priori, the diversity in the population is maintained without the need for computing crowding distances, which is one of the major costs in other multiobjective optimisation evolutionary algorithms, such as NSGA-II \cite{nsga2}.

MOEA/D has attracted much attention in the field of evolutionary multiobjective optimisation, and several modifications have been proposed, as surveyed in \cite{Zhou2011}. Some of the suggested improvements, for example, integration with the opposition-based learning \cite{Ma2014b}, Baldwinian learning \cite{Ma2014}, or end-user preference incorporation \cite{Ma2016} can be applied to the method proposed in this paper. The addition of these features is considered as future work.
	
\subsection{Multi-objective GAs for manufacturing scheduling}
	\label{sec:relWork:IPP}
	
One of the pioneering research related to industrial production planning using a multi-objective GA was described in \cite{Ishibuchi1998}. In that paper, a set of nondominated solutions was determined for a classic flowshop scheduling problem with three objectives, namely, the minimal makespan, the maximum tardiness and the total flowtime. A weighted sum of these three criteria was treated as a fitness value of each individual, but the weight values were randomly specified whenever a pair of the parent solutions was selected. Consequently, each point of the solution space was generated using a different weight vector. A local search was then applied for further improvement of those solutions. However, the considered problem was rather abstract and the considered plant and taskset sizes were limited \cite{Ishibuchi1998}.

Various real-world industrial scheduling problems were attempted to be solved with customised multi-objective GAs as well. In \cite{Li2009}, for example, a real-world manufacturing problem of a steel tube production was described as an extension of a classic Job-Shop Scheduling Problem with compatible resources (aka Flexible Job-Shop Scheduling Problem). A multi-objective GA was applied with two objectives: minimisation of the resources' idleness and waiting time of orders. In that paper, it stated explicitly that the prior research related to Job-Shop Scheduling Problems was impractical as being based on oversimplified models and assumptions. Nevertheless, that model is still inappropriate for the batch production problem considered in this paper. In particular, it is unable to select recipes or minimise the commodity surplus. Another interesting real-world manufacturing problem of textile batch dyeing scheduling was presented in \cite{Huynh2018}. In that problem, a batch is comprised of clothes of the same colour whose total weight does not exceed the capacity of the manufacturing resource. Again, that problem is different of the one analysed in this paper, as in the considered case the resources can produce only an exact weight of a given commodity. The total amount of manufactured commodities depends only on the recipes multisubset (i.e., a combination with repetitions) selected for manufacturing. Hence, a batching heuristics, as proposed in \cite{Huynh2018}, is not applicable, but a technique for optimisation of recipe multisubset selection is needed. GA was applied to such a problem in \cite{paintsOrig}, yet the optimisation was performed with typical multi-objective GAs (NSGA-II and MOEA/D). In particular, no linkage learning was performed in that approach. 
Note that in the research presented in this paper, we compare to both methods considered in \cite{paintsOrig} and both of these methods are outperformed by linkage learning GAs (namely MO-GOMEA and MO-P3) for the considered practical problem and for a wide set of benchmark problems.

More examples of applying multi-objective GAs to solve manufacture scheduling problems were surveyed in \cite{Gen2014}, including Job-Shop Scheduling Problems, Flexible Job-Shop Scheduling Problems, dispatching in flexible manufacturing systems and integrated process planning and scheduling. However, none of the papers reviewed there dealt with recipe multisubsets nor minimising the surplus of the manufactured commodities. Similarly, none of the reviewed papers used linkage learning to improve the performance of the applied GAs, as it is proposed in this paper.

	\section{Real-World Multi-Objective Bulk Commodity Production Problem Formulation}
	\label{sec:problemDef}
	
	In this section, the considered practical problem is firstly described and then formalised as a typical covering problem (CP) instance and its extension to multi-objectiveness.

	\subsection{Real-World Scenario Description}
	
	The considered scenario is based on a process of manufacturing, in which a certain amount of commodities is produced by combining supplies, ingredients or raw substances following a stored recipe. The main optimisation objective of this case study is to decrease the makespan of batch production. Depending on the selected multisubset (i.e., a combination with repetitions) of recipes, the time to produce a commodity may vary significantly, which influences the percentage of manufacturing time that is truly productive, known as Overall Equipment Effectiveness (OEE). 
	
	In the considered multi-objective bulk commodity production problem (MOB\\CPP), the recipes for each batch produce a certain amount of commodity. Consequently, to satisfy an order for a certain commodity, one or more recipes for producing such commodity have to be selected and allocated to resources. However, the sum of the commodity amount produced by any selection of recipes may be different from the order amount for that commodity. If a certain commodity cannot be produced in the required amount, some commodity surplus is expected. As the surplus storage can be expensive and larger surplus usually implies a higher cost of raw substances used in the production, additional optimisation objectives can be defined: not only the makespan but also the surpluses of each produced commodities have to be minimised. This observation leads to the conclusion that multi-objective optimisation techniques, as described earlier in this paper, can be applied. In particular, this problem can be viewed as a variant of the classic covering problem, as shown in the following subsection. 
	
	\subsection{Problem Formulation}
	
	The considered factory manufactures bulk commodities $c_j$, $j = 1, \ldots, m$. These commodities can be produced by executing $x_i$, $i=1,\ldots,n$, times some pre-defined manufacturing recipes $\gamma_i$ on the only resource $\pi$. The objective is to minimise the makespan 
	\begin{equation}
	\sum_{i=1}^{n}t_ix_i,
	\end{equation}
	where $t_i$ denotes the pre-defined execution time of recipe $\gamma_i$ subject to \begin{equation}\label{eq:constraints}
	\sum_{i=1}^{n}\delta_{i,j}x_i \geq o_j, 
	\end{equation}
	where $o_j$ denotes the ordered amount of commodity $c_j$, $\delta_{i,j}$ is the amount of commodity $c_j$ produced by recipe $\delta_i$. The problem defined in this way is a typical example of CP and, as such, is an instance of ILP and belongs to the NP-hard class \cite{Garey1990}.
	
	The first extension of the above problem is the possibility of multiple resources $\pi_i$ in the factory, each being capable of executing recipe $\gamma_i$. Hence the makespan minimisation objective can be rewritten as minimising 
	\begin{equation}
	\max_{\forall i \in \{1,\ldots,n\}} (t_ix_i)
	\end{equation}
	subject to the same constraints as provided in (\ref{eq:constraints}).
	
	The next modification is caused by the surplus storage cost in the factory and the cost of raw substances needed to produce the commodities, which force the factory to minimise not only the makespan, but also the surpluses of each commodity, i.e. to produce as little commodities as possible to satisfy the ordered amounts. Hence, the following $m$ objectives need to be added to the optimisation problem: $\forall j \in \{1,\ldots,m\}$ minimise 
	\begin{equation}
	\sum_{i=1}^{n}\delta_{i,j}x_i, 
	\end{equation}
	subject to the same constraints (\ref{eq:constraints}) as above.
	
	The number of instances of each recipe $\gamma_i$ can be viewed as being bounded by the ordered amount of commodities produced by this recipe, computed with equation 
	\begin{equation}\label{eq:ceil}
	\mu_i=\max_{\forall j \in \{1,\ldots,m\}} \left\lceil \frac{o_j}{\delta_{i,j}}\right\rceil.
	\end{equation}
	This upper-bound facilitates the binary encoding of the solutions for GA as described in the next section. The considered problem is a discrete combinatorial problem. As pointed in \cite{moSurvey}, if such problems are solved by conventional methods, the time required to solve them may increase exponentially. Therefore, the use of Multi-Objective Evolutionary Algorithms is justified.
	
	\section{Multi-Objective Parameter-less Population Pyramid for Solving MOBCPP}
	\label{sec:mop3}

	In this section, we describe the details, motivations and intuitions of the proposed Multi-Objective Parameter-less Population Pyramid (MOP3). In the first subsection, we discuss the solution encoding, while in the second one, we describe the proposed method.

	\subsection{Solution Encoding}
	\label{sec:mop3:encoding}
	
	As presented in the previous section, in the MOBCPP problem, the size of ordered amounts is known. However, the number of production tasks (\textit{jobs}) is not specified because each recipe may produce a different amount of paint.
	Let us consider an example with a single commodity. The ordered amount is $o_1 = 10$ units. There are two available recipes, $\gamma_1$ and $\gamma_2$, that produce $\delta_{1,1}=3$ and $\delta_{2,1}=5$ units of commodity $c_1$, respectively. Thus, based on equation (\ref{eq:ceil}), to satisfy order $o_1$, it is sufficient to execute $\mu_1=4$ jobs using $\gamma_1$ recipe or $\mu_2=2$ jobs that use recipe $\gamma_2$. Note that the solution to the problem instance considered in this example may be encoded as a 6-bit long binary string, where the first four bits refer to jobs executing recipe $\gamma_1$ and the last two bits refer to jobs that execute recipe $\gamma_2$.\par 
	
	Based on the above example, we may state that a solution to MOBCPP can be encoded as a binary string where a particular bit refers to a single job executing a particular recipe. In this paper, we order the bits in the following manner. First, we consider all bits that refer to the jobs producing the first commodity, then the bits that refer to the jobs producing the second commodity and so on. The jobs producing more than one commodity are located at the position suitable for the produced commodity with the lowest index. Among the bits that consider the production of a particular commodity, we first encode the bits referring to the minimum number of jobs using a recipe with the lowest index in the recipe list, then we encode the bits referring to the jobs using the recipe with the second-lowest index in the list and so on. The minimum number of jobs to satisfy the ordered amount of $o_i$, using recipe $\gamma_i$ that produces $\delta_{i,j}$ resource units is computed with equation (\ref{eq:ceil}).
	
	\begin{figure}[ht!]
		\centering%
		\includegraphics[width=0.9\textwidth]{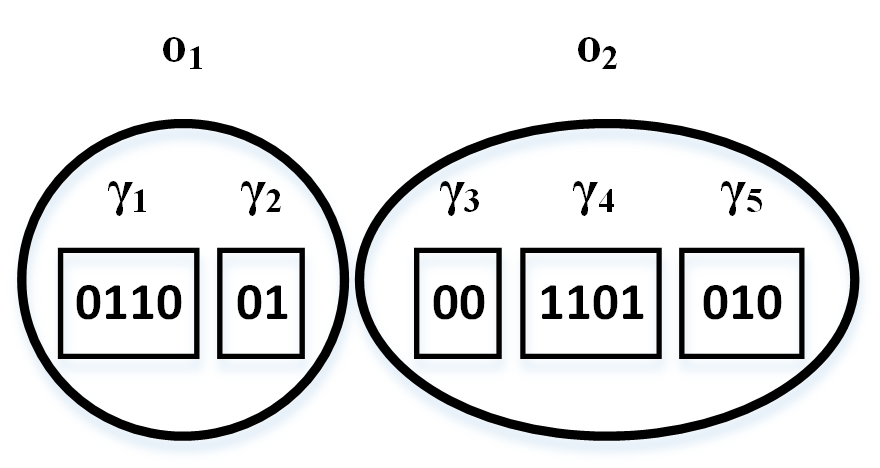}
		\caption{Solution encoding example}
		\label{fig:encoding:ex}
	\end{figure}
	
	Let us consider the following example. Two commodities with orders $o_1 = 10$ and $o_2 = 8$ units are considered. The first commodity may be produced with the use of $\gamma_1$ and $\gamma_2$ recipes that produce $\delta_{1,1}=3$ and $\delta_{2,1}=5$ of commodity units, respectively. The second commodity may be produced with the use of three recipes ($\gamma_3$, $\gamma_4$ and $\gamma_5$) that produce $\delta_{3,2}=5$, $\delta_{4,2}=2$ and $\delta_{5,2}=3$ units of this commodity. The encoding and the solution to this problem instance are presented in Fig. \ref{fig:encoding:ex}. First, the jobs that consider the order for the first commodity are considered. Among them, the first four bits refer to jobs using recipe $\gamma_1$, the next two refer to jobs using recipe $\gamma_2$. Respectively, for the order referring to the second commodity, the first two bits refer to recipe $\gamma_3$, the next four to recipe $\gamma_4$ and the last three to recipe $\gamma_5$.\par
	
	
	
	
	
	

	Note that Fig. \ref{fig:encoding:ex} presents a feasible solution, i.e., the one that produces enough commodities. However, the solution encoded in the manner proposed above may be infeasible. Moreover, it may exceed the order size. Therefore, to fix the above issues, we propose a genotype repair algorithm presented in Pseudocode \ref{alg:imAndSdsnc}. Any genotype that has been updated by the proposed algorithm is feasible and it does not use more jobs for a particular recipe than necessary (i.e., all jobs that may be abandoned without violating the order constraint will be removed). The proposed solution encoding is not unique as two or more different genotypes can encode the same solution. The genotype repair algorithm works as follows. For each gene (that corresponds to a particular recipe), the list of orders produced by the corresponding recipe is gathered. If, for at least one order, we need to increase the production amount, then we set the gene value on $1$. If, for all orders, we can resign from using the recipe without violating the order size, then we set the gene value on $0$. Otherwise, the gene value remains unmodified.

	\begin{algorithm}
		\caption{Genotype repair algorithm}
		\begin{algorithmic}[1]
			\Procedure{Repair}{Genotype}
			\State ResGenotype = Genotype
			\For{$Gene \gets 1$ \textbf{to} length($ResGenotype$)}
			
			\State $OrdersIndexes \gets$ GetOrdersForGene($ResGenotype, Gene$)
			\State $decision = -1$
			\For{$i \gets 1$ \textbf{to} length($OrdersIndexes$)}
			\State $Order \gets OrdersIndexes[i]$
			\State $OrderToDo \gets $ GetOrderSize($Order$)
			\State $OrderPlanProd \gets $ GetPlanProd($Order,ResGenotype$)
			\State $RecipeAmount \gets$ GetRecipeAmountForGene($Order,Gene$)
			
			\If {$OrderToDo > OrderPlanProd$} 
			\State $decision = 1$
			\EndIf
			
			\If {$OrderToDo > OrderPlanProd - RecipeAmount$ \textbf{and} $decision = -1$} 
			\State $decision = 0$
			\EndIf
			\EndFor
			
			\If {$decision = 1$} 
			\State $ResGenotype[Gene] = 1$
			\EndIf
			
			\If {$decision = -1$} 
			\State $ResGenotype[Gene] = 0$
			\EndIf
			
			\EndFor		
			
			\State \Return{ResGenotype}
			
			\EndProcedure
		\end{algorithmic}
		\label{alg:imAndSdsnc}
	\end{algorithm}

	\subsection{Multi-Objective Parameter-less Population Pyramid}
	\label{sec:mop3:mmethod}
	
	In this section, we present the motivations behind the proposed Multi-Objective Parameter-less Population Pyramid (MO-P3) and its description. The main reason for using P3 as a research starting point for proposing a method dedicated to solving a practical multi-objective problem considered in this paper are enumerated below.\par

	\textbf{Motivation 1}. P3 is effective in solving problems with overlapping building blocks (see Section \ref{sec:relWork:linkageDiversity}). The relations between building blocks (overlapping) are typical for real-life problems \cite{watsonHiffPPSNfirst,OverlappingSimon}. P3 has also been found more effective than LTGA and DSMGA-II when applied to a single-objective practical problem \cite{steinerTrees} (to the best of our knowledge, the results presented in \cite{steinerTrees} are the only comparison between P3, LTGA and DSMGA-II based on practical problem instances). Thus, it is reasonable to assume that a P3-based method dedicated to solving a practical multi-objective problem may be found more effective than MO-GOMEA (that is LTGA-based).\par

	\textbf{Motivation 2}. As explained in Section \ref{sec:relWork:moGomea}, MO-GOMEA clusters its population concerning the objective space. MO-GOMEA maintains separate linkages for each Pareto front cluster. This kind of multi-objective problem decomposition is the key feature of MO-GOMEA and one of the reasons for its high effectiveness. MO-GOMEA is based on the idea of LTGA, which maintains a single linkage at a time. On the other hand, P3 maintains many linkages (one per pyramid level). As explained in Section \ref{sec:relWork:dsmMethods:p3}, this linkage diversity is likely to be the reason for the high effectiveness of P3 in solving the heavily overlapping problems. Maintaining many different linkages facilitates identifying different blocks (groups of gene indexes). The same feature is required when solving multi-objective problems. Thus, a P3-based method may be highly effective in solving multi-objective problems even without problem decomposition performed by MO-GOMEA.\par
	
	Considering the above motivations, we propose a Multi-Objective Parameter-less Population Pyramid (MO-P3) that includes the following mechanisms. MO-P3 uses the elitist archive in the same way as MO-GOMEA (see Section \ref{sec:relWork:moGomea}). In the original P3, each individual that climbs up the pyramid optimises a single-objective problem. Therefore, to adjust MO-P3 to multi-objective optimisation, when a new iteration of MO-P3 starts, a normalised weight vector is chosen to transform a multi-objective problem into a single-objective one. Thus, during its climbing, each individual optimises a single-objective problem. However, due to different weights, it shall climb to reach a different part of the Pareto front.\par

	\begin{algorithm}
		\caption{The general MO-P3 overview}
		\begin{algorithmic}[1]
			\Procedure{MO-P3}{}
			
			\State $levels \gets $ \{ CreateNewEmptyPop()\} \Comment{initialization}
			
			\While{$\neg stopCondition$}
				\State $weightVec \gets $ GetWeightVector();
				\State $newInd \gets $ GetRandomIndividual();
				\State $newInd \gets $ FIHC($newInd, weightVec$);
				\If {$newInd$ $\neg$ exist in $levels$}
					\State InsertIndividualOnLevel($levels[0]$,$newInd$);
				\EndIf
				
				\For {\textbf{each }$level \in levels$}
					\State $IndImpr \gets $ ImproveIndWithLevel($level$, $newInd$, $weightVec$);
					\If {$IndImpr \neq newInd$ }
						\State $newInd \gets IndImpr$
						\If {$newInd$ $\neg$ exist in $levels$}
							\State $nextLevel \gets$ GetNextLevel($level$);
							\If {$nextLevel = empty$ }
								\State $nextLevel \gets $ \{ CreateNewEmptyPop()\} 
								\State AddNewTopLevel($levels, nextLevel$);								
							\EndIf							
							\State InsertIndividualOnLevel($nextLevel$,$newInd$);
						\EndIf						
					\EndIf
				\EndFor
				
			\EndWhile

			\EndProcedure		
		\end{algorithmic}
		\label{alg:mop3Overview}
	\end{algorithm}

	The general overview of MO-P3 work is presented in Pseudocode \ref{alg:mop3Overview}. For each new individual, the weight vector is chosen to direct the search in one of the parts of the Pareto front (lines 4 and 5). The new individual is improved by FIHC and added to the pyramid (lines 6-9). Then, the new individual climbs up the pyramid. Note that the chosen weight vector is used during the whole iteration. The new individual may cross with any other individual that was added to the pyramid. However, any time the new individual is mixed with individuals on the given pyramid level (line 11), the linkage gathered for this level is employed, and the search is directed by the weight vector selected at the beginning of the iteration (line 4).\par
	
	The way of choosing the weight vector at the beginning of each MO-P3 iteration is crucial. It shall push the search to focus on different parts of the Pareto front. However, if the search is too heavily biased towards some parts of the Pareto front, the linkage diversity offered by the pyramid-like population structure may not be sufficient. In consequence, the linkage gathered by MO-P3 may become useful to optimise only some parts of the Pareto front. If so, the quality of solutions referring to other Pareto front parts (those for which the linkage stored by MO-P3 is not useful) may be low. In this paper, we consider two different strategies of choosing the weight vector. In the first one, the weight vector is chosen randomly. MO-P3 using this technique will be denoted as MO-P3-Random. The second strategy of choosing the weight vector is presented in Pseudocode \ref{alg:smartWeightVector} and denoted as MO-P3-Smart.

	\begin{algorithm}
		\caption{Smart strategy of choosing the two-dimensional weight vector}
		\begin{algorithmic}[1]
			\Procedure{GetWeightVector}{ElitistArchive}
			
			\State $EApoints \gets empty$
			\For {\textbf{each} \textit{sol} in \textit{ElitistArchive}}
			\State $sum = sol.FirstObjNormalised + sol.SecondObjNormalised$
			\State $newPoint.FirstWeight = sol.FirstObjNormalised / sum$
			\State $newPoint.SecondWeight = sol.SecondObjNormalised / sum$
			\State $EApoints \gets newPoint$
			\EndFor \text{\textbf{ each}}
			
			\State $EApoints \gets$ Sort($EApoints$)
			
			\State $interv_1 \gets$ GetRandomInt($1, $ SizeOf($EApoints$)$-1$)
			\State $interv_2 \gets$ GetRandomInt($1, $ SizeOf($EApoints$)$-1$)
			
			\State $length_1 \gets $ GetEuclDist($EApoints.At(inter_1), EApoints.At(inter_1 + 1)$)
			\State $length_2 \gets $ GetEuclDist($EApoints.At(inter_2), EApoints.At(inter_2 + 1)$)

			\If {$length_1 > length_1$}
			\State $intervChosen \gets interv_1$
			\Else
			\State $intervChosen \gets interv_2$
			\EndIf
			
			\State $IntervalStart = EApoints.At(intervChosen).FirstObj$
			\State $IntervalEnd = EApoints.At(intervChosen).SecondObj$
			
			\State $weightVec.First = $ GetRandomReal($IntervalStart, IntervalEnd$)
			\State $weightVec.Second = 1 - weightVec.First$

			\State \Return{weightVec}
			
			\EndProcedure		
		\end{algorithmic}
		\label{alg:smartWeightVector}
	\end{algorithm}
	
	In the \textit{smart} strategy of choosing the weight vector, each solution in the elitist archive is transformed into a weight vector. Such an array of weight vectors is then sorted concerning the weight corresponding to the first objective. After sorting an array of weights, the vectors may be interpreted as an array of the crowding distances \cite{nsga2}. Then, the tournament of size two is used to choose the interval that refers to the higher value of the crowding distance. The weight vector is randomly chosen from the interval returned by the tournament.\par
	
	The motivation behind the \textit{smart} strategy of choosing the weight vector is as follows. MO-P3-Smart is expected to push the search towards these regions of the Pareto front that are poorly represented. On the other hand, such a bias may cause the linkage to be useful to optimise only some parts of the Pareto front. In this case, the overall Pareto front quality may decrease significantly.\par
	
	In this section, we have proposed a multi-objective method dedicated to solving the MOBCPP problem. We propose both: the problem-dedicated solution encoding and the new method. MO-P3 is based on the P3 method. The key modification is the choice of weight vector for further optimisation of a new individual during the climb. We propose two strategies for choosing the weight vector: Random and Smart. In the subsequent sections, we show that our proposition is more effective than the competing methods in solving both: the MOBCPP problem and the typical benchmarks employed in multi-objective discrete optimisation.

	\section{Experiments on MOBCPP}
	\label{sec:exp:paints}
	In this section, we present the results obtained for the MOBCPP problem. The objective of the experiments was to compare the effectiveness of the proposed MO-P3 with other methods dedicated to multi-objective optimisation on the base of the MOBCPP problem. The rest of this section is organised as follows. In the first subsection, we present the experiment setup. In Section \ref{sec:exp:paints:strategiesComparison}, the two considered MO-P3 versions are compared. In the third subsection, we analyse the fitness evaluation number (FFE) and computation time ratio. Finally, in the last subsection, we compare MO-P3 with MO-GOMEA, MOEA/D and NSGA-II.

	\subsection{Experiments Setup}
	\label{sec:exp:paints:setup}
	
	In this paper, we consider $27$ different MOBCPP problem instances. Each instance is related to a real-life configuration that takes place or may take place in practice. We consider two groups of test cases. In the first group (16 test cases), we consider one production hall, with multiple machines. In these scenarios, each job can be executed on any machine (more information about these scenarios is given in Table \ref{tab:testCases}). In the second group of test cases (11 test cases), we consider scenarios with multiple production halls. In each hall, we can produce only a subgroup of paints (more information is given in Table \ref{tab:testCasesMulti}). Note that even for a relatively low number of resources, the number of available encodings is large. Since MOBCPP is NP-hard (see Section \ref{sec:problemDef}), the considered test cases may be found difficult to solve. Each experiment has been repeated 50 times.\par
	
	\begin{table}
		\caption{The parameters of the single-hall test cases}
		\centering%
		\label{tab:testCases}
		\begin{tabular}{ccc}
			\hline
			\textbf{Parameter} & \textbf{Min} & \textbf{Max} \\
			\hline
			\textbf{Production halls} & \multicolumn{2}{c}{1} \\
			\textbf{Resources} & 2 & 3 \\
			\textbf{Commodities} & 6 & 20 \\
			\textbf{Recipes} & 12 & 60 \\
			\textbf{Genotype length} & 46 & 746 \\
			\textbf{Encodable solutions} & $7.03 \cdot 10^{13}$ & $3.70 \cdot 10^{224}$\\
			\hline
		\end{tabular}
	\end{table}

	\begin{table}
		\caption{The parameters of the multi-hall test cases}
		\centering%
		\label{tab:testCasesMulti}
		\begin{tabular}{ccc}
			\hline
			\textbf{Parameter} & \textbf{Min} & \textbf{Max} \\
			\hline
			\textbf{Production halls} & 2 & 12 \\
			\textbf{Resources} & 2 & 24 \\
			\textbf{Commodities} & 6 & 72 \\
			\textbf{Recipes} & 12 & 144 \\
			\textbf{Genotype length} & 92 & 552 \\
			\textbf{Encodable solutions} & $10^{27}$ & $10^{162}$\\
			\hline
		\end{tabular}
	\end{table}

	We consider two MO-P3 versions that employ two different strategies for weight vector initialisation (see Section \ref{sec:mop3:mmethod}). Depending on the employed strategy, they will be denoted as MO-P3-Random and MO-P3-Smart, respectively. 
	
	We use three competing methods: Non-dominated Sorting Genetic Algorithm II (NSGA-II) \cite{nsga2}, Multi-Objective Evolutionary Algorithm based on Decomposition (MOEA/D) \cite{moead} and Multi-objective Gene-pool Optimal Mixing Evolutionary Algorithm (MO-GOMEA) \cite{MoGomeaSwarm}. NSGA-II has been selected as it is commonly employed as a baseline in the multi-objective optimisation domain. Similarly to \cite{MoGomeaGecco}, we use bit-flipping mutation with probability $1/l$, where $l$ is the genotype length, the probability of crossover is 0.9. To make NSGA-II independent of the gene order, we use the uniform crossover. Finally, we consider the population sizes of 25, 50, 100, 200, 400 individuals, the same as in \cite{MoGomeaGecco,MoGomeaSwarm}. MOEA/D is frequently used as a research starting point and as a baseline \cite{MoGomeaSwarm,Zhou2011}. Similarly to NSGA-II, MOEA/D requires specification of the population size. We consider the same population sizes as in the NSGA-II case. MO-GOMEA is a state-of-the-art method in the multi-objective optimisation domain that has been reported to significantly outperform NSGA-II \cite{MoGomeaGecco,MoGomeaSwarm} and MOEA/D \cite{MoGomeaSwarm}. Note that MO-GOMEA and the proposed MO-P3 are parameter-less methods. Thus, no tuning is necessary. This feature makes these methods particularly useful for practical implementations. We have abandoned the comparison with the Multi-objective Hierarchical Bayesian Optimization Algorithm \cite{mohBOA} because it was significantly outperformed by MO-GOMEA \cite{MoGomeaGecco,MoGomeaSwarm}.\par
	
	For the considered methods, we use the source codes published by their authors\footnote{\url{http://www.iitk.ac.in/kangal/codes.shtml} for NSGA-II, the source code used in \cite{MoGomeaSwarm} for MO-GOMEA, the source code given in \cite{P3Original} for P3 and \url{https://github.com/ZhenkunWang} for MOEA/D}. All sources have been merged on the problem definition level in one project that is available at \url{https://github.com/przewooz/moP3}. Additionally, with the source code, all settings files and the detailed results of all experiments are provided.\par
	
	As the stop condition, we use the fitness function evaluation number (FFE). This choice is motivated by the significant amount of computation time consumed by the fitness value computation. The computation budget has been set to 25 million fitness evaluations.\par

	As a quality measure, we use the Inverted Generational Distance (IGD). IGD is defined as 
	\begin{equation}
	\label{eq:igd}
	D_{\mathcal{P}_F\to \mathcal{S}}(\mathcal{S}) = \frac{1}{|\mathcal{P}_F|} \sum_{\substack{f^0 \in \mathcal{P}_F}} \min_{x \in \mathcal{S}}{\{d(f(x), f^0) \}},
	\end{equation}
	where $\mathcal{P}_F$ is the Pareto-optimal front, $\mathcal{S}$ is the final front proposed by the optimiser and $d(\cdot,\cdot)$ is the Euclidean distance. IGD is an average distance from each point in $\mathcal{P}_F$ to the nearest point in $\mathcal{S}$. The quality of the proposed front $\mathcal{S}$ is inversely proportional to the IGD value. The optimal IGD value is $0$, which means that $\mathcal{S}$ covers $\mathcal{P}_F$. We can also compute the average distance from each point in $\mathcal{S}$ to the nearest point in $\mathcal{P}_F$. Such a measure is called the Generational Distance (GD) \cite{MoGomeaSwarm}. The advantage of IGD over GD is that IGD value is optimal if and only if  $\mathcal{S}$ covers the whole $\mathcal{P}_F$. Oppositely, GD value is optimal if $\mathcal{S}$ is a subset of $\mathcal{P}_F$ \cite{MoGomeaGecco}. Therefore, we favour IGD over GD.\par
	
	The optimal Pareto front must be known to compute IGD. Unfortunately, the considered test cases are based on practice and the optimal Pareto front is not known. To overcome this issue, we construct a pseudo-optimal Pareto front in the following way. For each test case, we consider all $\mathcal{S}$ fronts proposed by every method in every run. From this set of points, we choose only non-dominated ones. The pseudo-optimal Pareto front obtained this way may not be optimal. Nevertheless, all considered $\mathcal{S}$ fronts only contain points that are the part of pseudo-optimal Pareto front or that are dominated by points from pseudo-optimal Pareto front. The same procedure of pseudo-optimal Pareto front creation has been applied in \cite{MoGomeaGecco}.

	\subsection{The Comparison between MO-P3 with Random and Smart Strategies}
	\label{sec:exp:paints:strategiesComparison}
	
	To compare the performance of MO-P3-Random and MO-P3-Smart, we consider the median IGD value that describes the quality of the proposed Pareto front and the median FFE number necessary to obtain the final solution. To check the statistical significance of the differences, we use the Wilcoxon signed-rank test with a typical 5\% significance level. The summarised results are presented in Table \ref{tab:radnomVsSmart}.\par

	\begin{table}
		\caption{The effectiveness comparison between MO-P3 employing Random and Smart strategies}
		\centering%
		\label{tab:radnomVsSmart}
		\begin{tabular}{ccccc}
			\hline
			\textbf{Test-case type} & & \textbf{Random} & \textbf{equal} & \textbf{Smart} \\
			\hline
			\multirow{2}{*}{\textbf{Single-hall}} & \textbf{IGD} & 7 & 9 & 0 \\
			 & \textbf{Median FFE until final solution} & 8 & 7 & 1 \\
			 \multirow{2}{*}{\textbf{Multi-hall}} & \textbf{IGD} & 0 & 11 & 0 \\
			 & \textbf{Median FFE until final solution} & 0 & 11 & 0 \\
			\hline
		\end{tabular}
	\end{table} 

	For test cases with a single production hall, MO-P3 with the \textit{random} strategy has outperformed MO-P3-Smart for 7 test cases (out of 16) and has never been found inferior. Moreover, MO-P3-Random has also been faster to find the solution in 50\% of the cases and found slower for only one test case.\par 
	
	The \textit{smart} strategy chooses the weight vector at each MO-P3 iteration in a way that shall force the method to obtain a more diverse Pareto front. Thus, it may be found surprising that MO-P3-Smart has been outperformed by MO-P3-Random. However, as stressed in Section \ref{sec:mop3:mmethod}, the \textit{smart} strategy may bias the method towards some solution space regions. If this happens, the linkage gathered and utilised by MO-P3 may become useful only for improving some parts of the Pareto front. As a consequence, the overall Pareto front quality will drop. Note that the drop or lack of linkage diversity may cause a method to become ineffective \cite{3lo}.\par
	
	For the test cases considering many production halls, both strategies report results of equal quality. The differences in median FFE necessary for reaching the best result are not statistically significant. Such results may be found surprising when compared to those obtained for a single production hall. The reasonable explanation of this fact is as follows. When we consider multiple halls and each hall produces only a subgroup of paints, then the key issue is to successfully find the appropriate linkage that divides a genotype into subparts responsible for production on each of the halls. For DSM-using methods (like P3, or MO-GOMEA), such a task may be easy for some problem types \cite{linkageQuality}, and hard for the other. If for these test cases, the key to finding a high-quality Pareto front is to find a high-quality linkage that divides a genotype into the appropriate parts, then the multi-level population structure of MO-P3 is the key to solve these problem instances. The other MO-P3 features that cause the domination of MO-P3-Random over the MO-P3-Smart in the case of single-hall test cases do not seem to significantly influence the results for test cases considering many production halls.\par

	\begin{figure}
		\includegraphics[width=\linewidth]{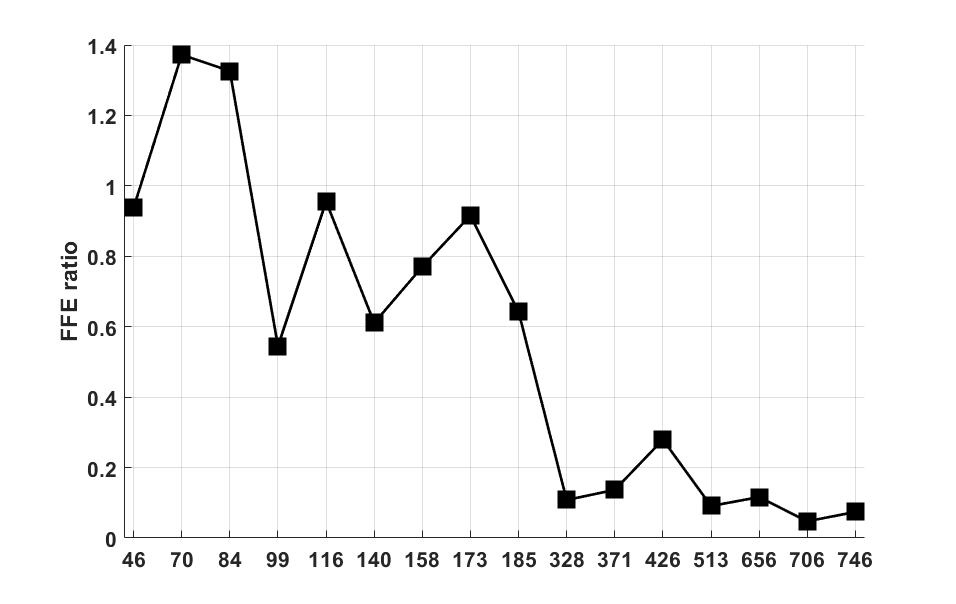}
		\caption{The $FFE_{Random}/FFE_{Smart}$ ratio of FFE spent on reaching the final result by MO-P3-Random and MO-P3-Smart for test cases with a single production hall}. Horizontal axis: problem size
		\label{fig:ffeRationSmartvSRandom}
	\end{figure}

	In Figure \ref{fig:ffeRationSmartvSRandom}, we show the $FFE_{Random} / FFE_{Smart}$ ratio for test cases with a single production hall. The $FFE_{Strategy}$ is the median FFE necessary for finding the final solution. The figure shows which method has been faster depending on the number of genes necessary to encode the problem solution. We abandon such a comparison for the multi-hall test cases because there are no statistically significant differences in the FFE number necessary to find the final solution between MO-P3-Random and MO-P3-Smart. If the value of the ratio is below $1$, then MO-P3-Random is faster, if it is higher, then the situation is the opposite. Note that the longer is the genotype, the faster is MO-P3-Random when compared to MO-P3-Smart. Such observation is coherent with the conclusion that the likely reason for the low effectiveness of MO-P3-Smart is the loss of linkage diversity. The lower is the number of genes, the less important is the quality of linkage \cite{linkageQuality}. In other words, the chance for successful crossing is inversely proportional to the genotype length (see Section \ref{sec:exp:benchmarks}). That is why MO-P3-Smart is almost equally fast to find the final solution for the genotypes of the length not exceeding 200 genes (for two test cases, it is even faster than MO-P3-Random). However, for longer genotypes, MO-P3-Random is up to twenty times faster (with equal or better results quality). The reasonable explanation of this observation is that MO-P3-Random posses linkage that is good enough to optimise any part of the Pareto front, while MO-P3-Smart does not due to the bias caused by the \textit{smart} strategy. The situation in which precisely constructed algorithms (or their parts) are outperformed by their random-based competitors is rather rare and may be found as a phenomenon. However, in the literature of the field, we may point the similar cases \cite{linkageLearningIsBad2}.\par

	Since MO-P3-Random outperforms MO-P3-Smart, in the latter parts of this paper, we will consider only MO-P3 that employs the \textit{random} strategy only. Thus, whenever we refer to MO-P3, we mean MO-P3-Random.

	\subsection{FFE and Computation Time Ratio Comparison}
	\label{sec:exp:paints:ffeTimeRatio}
	
\begin{figure}
		\includegraphics[width=\linewidth]{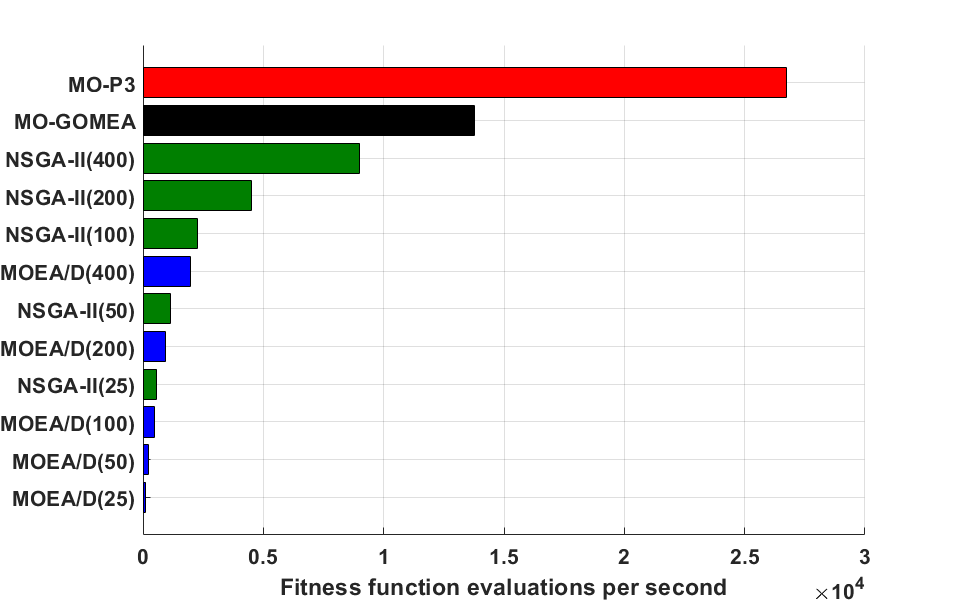}
		\caption{Median $FFE/ComputationTime$ ratio per method for all considered test cases}
		\label{fig:plotFfeRatio}
	\end{figure}

	Figure \ref{fig:plotFfeRatio} presents the median fitness function evaluation number per second ratio. All MOBCPP test cases have been considered. The values have been measured for short 10-minute runs performed on PowerEdge R430 Dell server Intel Xeon E5-2670 2.3 GHz 64GB RAM with Windows 2012 Server 64-bit installed. To assure the precision of computation time measurement, the number of computation processes has always been one fewer than a number of available CPU nodes. All experiments have been executed in a single thread without any other resource-consuming processes running. Such an experiment setup seems reliable for experiments using a time-based stop condition. Similar experiment setup may be found in \cite{muppets,3lo,muppetsActive}\par
	
	As stated in Subsection \ref{sec:exp:paints:setup}, for the considered test problem, the significant amount of computation resources is spent only on fitness value computation. Nevertheless, the $FFE/ComputationTime$ ratio comparison is important as it shows which method is faster.
	
	As presented in Figure \ref{fig:plotFfeRatio}, MO-P3 is significantly faster than any other considered method. The statistical significance of median differences has been confirmed by the Wilcoxon signed-rank test. For the null hypothesis that the median $FFE/ComputationTime$ ratio is equal to the median of any other method, the \textit{p}-value has not been higher than $10^{-56}$. NSGA-II and MOEA/D results are dependent on the population size. Such an observation is expected since the smaller the population size is, the more likely the population is to stuck. When the population is stuck, the method frequently requires fitness computation for the same genotypes. If this happens, the fitness may be recomputed or recovered from the caching buffer that stores the fitness values for some of the already considered genotypes. In the considered experiments, all methods have been joined on the problem definition level and the fitness is not recomputed if an individual remains unchanged. Therefore, the $FFE/ComputationTime$ ratio for NSGA-II and MOEA/D with low population size values is low. Note that fitness caching may lead to the following consequences. If the method is stuck and tends to consider the same and small subset of encodable genotypes, the $FFE/ComputationTime$ ratio may become so low, that the computation resources spent on other method activities than the fitness computation will become so significant that FFE will not be a fair and reliable measure. A more detailed analysis of this phenomenon may be found in \cite{fitnessCaching,PEACh,PEAChWPlfl}. Due to the very low $FFE/ComputationTime$ ratio values obtained for NSGA-II and MOEA/D, the considered population size for these methods is 400 individuals hereafter.

	\subsection{The Comparison between MO-P3 and the Competing Methods}
	\label{sec:exp:paints:moP3moGomea}

	\begin{figure}
		\includegraphics[width=\linewidth]{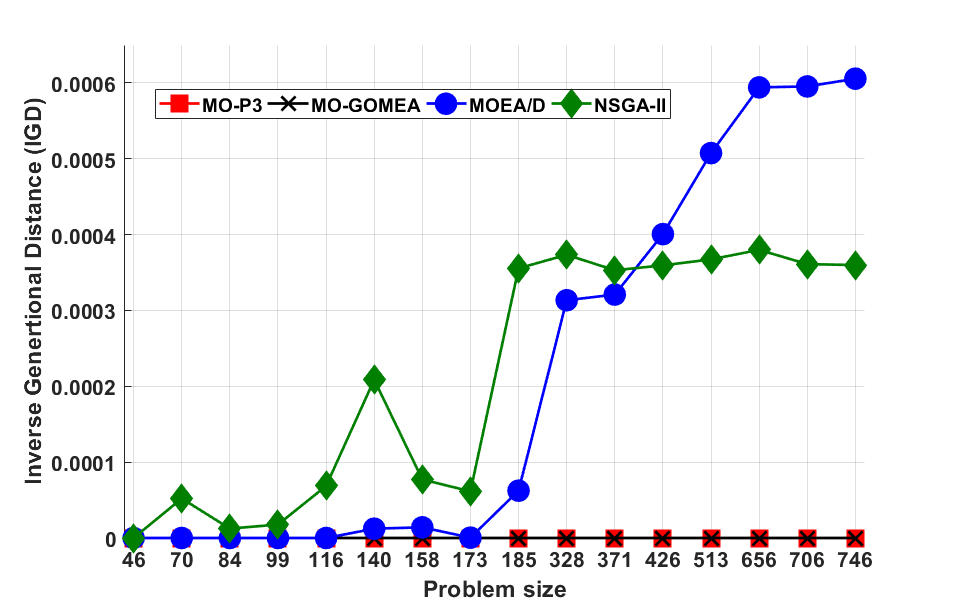}
		\caption{The IGD-based comparison of considered methods for test cases using a single hall}
		\label{fig:paintsSingleHall}
	\end{figure}

	\begin{figure}
		\includegraphics[width=\linewidth]{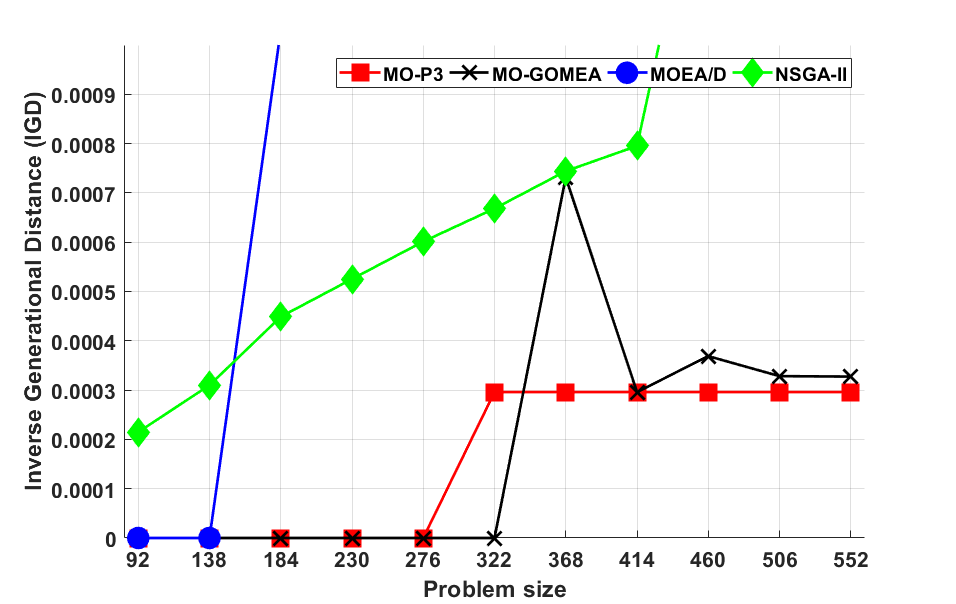}
		\caption{The IGD-based comparison of considered methods for multi-hall test cases}
		\label{fig:paintsMultiHall}
	\end{figure}
	
	The IGD comparison between MO-P3 and the rest of the considered methods is presented in figures \ref{fig:paintsSingleHall} and \ref{fig:paintsMultiHall}. MO-P3 outperforms NSGA-II and MOEA/D for both types of test cases. Such a result is expected since neither NSGA-II nor MOEA/D uses linkage information. Therefore, these methods are not capable of recognising the nature of the problem and do not use this knowledge to improve the effectiveness of the optimisation process. Thus, in the latter part of this subsection, we compare MO-P3 with MO-GOMEA that employs linkage learning, the same as MO-P3 is parameter-less and has been proposed recently to solve multi-objective problems effectively.\par

	\begin{figure}
		\includegraphics[width=\linewidth]{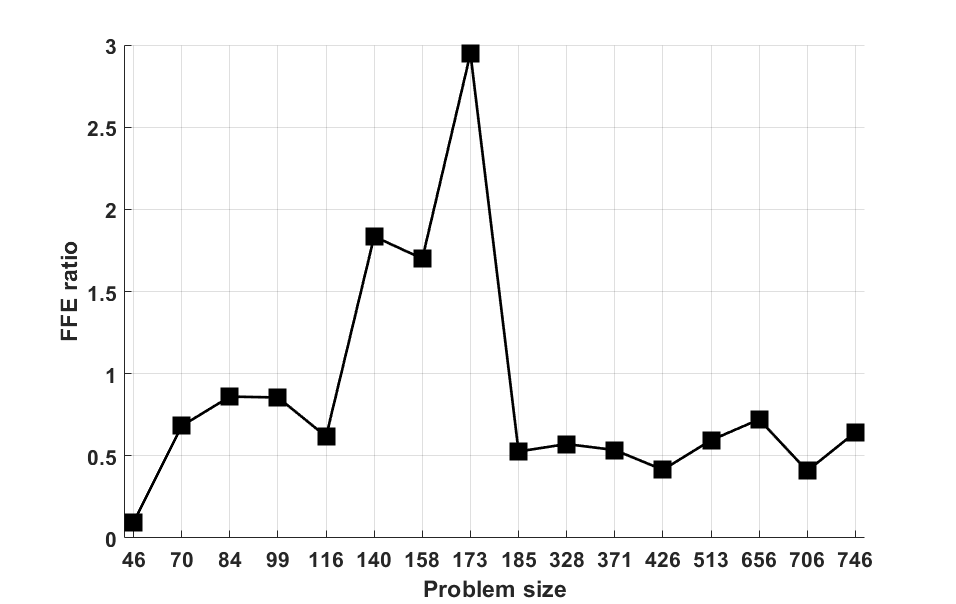}
		\caption{The ratio of FFE spent on reaching the final result by MO-P3 and MO-GOMEA for test cases with a single production hall}
		\label{fig:ffeRatioMOP3vsMOGOMEA}
	\end{figure}

	For all the test cases considering a single production hall, the median IGD values obtained by MO-P3 and MO-GOMEA have been equal. However, MO-P3 has been significantly faster in finding the solution in most of the runs. The Wilcoxon signed-rank test with the 5\% significance level has confirmed that these differences are statistically significant for 7 test cases. For another 7 test cases, the differences have not been meaningful and MO-GOMEA has been faster for two test cases. Figure \ref{fig:ffeRatioMOP3vsMOGOMEA} shows the $FFE_{MOP3} / FFE_{MO-GOMEA}$ ratio (similarly to the ratio shown in Figure \ref{fig:ffeRationSmartvSRandom}). For most of the considered test cases, MO-P3 is about two times faster than MO-GOMEA. However, the FFE ratio presented in Figure \ref{fig:ffeRatioMOP3vsMOGOMEA} does not seem related to the genotype length. Such an observation has been expected as both the compared methods gather linkage and try to make it diverse. MO-GOMEA supports different linkages necessary to optimise different Pareto front parts by individuals clustering, while MO-P3 uses a pyramid-like population structure.\par

	The situation is different for multi-hall test cases. For most of the test cases, both methods report similar IGD, but MO-P3 outperforms MO-GOMEA for four test cases and is outperformed only in one of them. The difference between single- and multi-hall test cases is that to successfully solve them, a method has to precisely decompose the problem into parts referring to different production halls. MO-P3 maintains a larger number of linkage sets. Thus it is more likely that some of these linkages will be precise enough to allow for successful exchange for some of the halls (see Section \ref{sec:relWork:linkageDiversityExample}). Therefore, for a relatively low number of halls, both methods report results of equal quality (for one of them MO-GOMEA even outperforms MO-P3). However, when the number of halls increases, MO-P3 outperforms MO-GOMEA. These differences are statistically significant. Moreover, MO-P3 has been faster than MO-GOMEA in finding the final result (in terms of FFEs) for 7 test cases of 11. For other test cases, the differences have been not statistically significant.\par
	
	For the MOBCPP problem, MO-P3 and MO-GOMEA yield results of similar quality. However, MO-P3 has outperformed MO-GOMEA for four test cases when many production halls are considered and has been outperformed by MO-GOMEA for one test case only. Since these results are statistically significant, we may state that MO-P3 outperforms MO-GOMEA for the MOBCPP problem. An important advantage of MO-P3 over MO-GOMEA is that MO-P3 requires fewer FFEs to reach the final result. This situation takes place for 11 out of 27 considered test cases, and only for two test cases, MO-GOMEA is better. If we consider the fact that MO-P3 performs about two times more FFEs per second than MO-GOMEA (see Figure \ref{fig:plotFfeRatio}), we can state that MO-P3 is better in solving MOBCPP because it reports the results of slightly higher quality and is significantly faster in reaching these results in terms of FFEs and computation time.\par
	
	In this section, we have shown that the proposed MO-P3 outperforms NSGA-II and MOEA/D for the considered practical problem. It has been also demonstrated that it performs better and significantly faster than MO-GOMEA. The intuition that in multi-objective optimisation the P3-based methods may be significantly faster than LTGA-based (GOMEA-based) in solving practical problem instances has been confirmed. To get the full view on MO-P3 effectiveness, we report the comparison based on popular theoretical benchmarks in the next section.

	\section{Experiments on Benchmark Problems}
	\label{sec:exp:benchmarks}
	In this section, we compare MO-P3 with MO-GOMEA, MOEA/D, and NSGA-II using various theoretical benchmarks. The objective of this comparison is to evaluate the overall MO-P3 effectiveness in solving assorted multi-objective problems. As a baseline, we use NSGA-II and MOEA/D. Similarly to the results presented in \cite{MoGomeaGecco,MoGomeaSwarm}, MOEA/D and NSGA-II were outperformed by MO-GOMEA. In this section, we show that except for one benchmark problem (multi-objective knapsack), MO-P3 performs even better than MO-GOMEA.

	\subsection{Benchmark Problems}
	
	In this section, we present the considered benchmark problems. We use the same benchmarks as in \cite{MoGomeaGecco,MoGomeaSwarm}. For Multi-objective weighted MAXCUT and Multi-objective knapsack, we have adopted the implementation from \cite{MoGomeaSwarm} and developed our implementations for \textit{Zeromax-Onemax}, \textit{Trap5-Inverse Trap5} and \textit{Leading Ones Trailing Zeros} (LOTZ).

	\subsubsection{Zeromax-onemax}
	
	The objectives for \textit{Zeromax-onemax} problem are defined as 
	\begin{equation}
	\label{eq:probDef:oneMaxZeroMax}
	f_{Onemax}(u) = u;  f_{Zeromax}(u) = l-u,
	\end{equation} 
	where $l$ is the genotype length and $u$ is the \textit{unitation} (see Section \ref{sec:relWork:linkageDiversity}). Thus, $f_{Onemax}$ maximises the number of ones in the genotype, while $f_{Zeromax}$ maximises the number of zeros. The optimal Pareto front $\mathcal{P}_F$ contains $l+1$ points. Many solutions can refer to a single point on $\mathcal{P}_F$ except for the two extreme points that contain only ones and only zeros. Thus, it is potentially harder to find extreme parts of $\mathcal{P}_F$ than the extreme regions \cite{mohBOA}. Note that every encodable solution lies on $\mathcal{P}_F$.

	\subsubsection{Trap5-Inverse Trap5}
	
	Deceptive function of unitation~\cite{decFunc} has been introduced in Section \ref{sec:relWork:linkageDiversity} in formula (\ref{eq:dec3}). The inverse deceptive function of unitation is defined as
	\begin{equation}
	\label{eq:dec3inverse}
	\mathit{dec_{inverse}(u)}=
	\begin{cases}
	u - 1 & \text{if } u > 0\\
	k & \text{if } u = 0
	\end{cases},
	\end{equation}
	where $u$ is a sum of gene values (so called \textit{unitation}) and $k$ is a deceptive function size.
	
	In this paper, we use $k=5$, which is the same setting as used in \cite{MoGomeaGecco,MoGomeaSwarm}. The \textit{Trap5-Inverse Trap5} problem is a concatenation of order-5 deceptive blocks. The first objective refers to the trap-5 function and maximises the blocks built from ones, while the second objective maximises the number of blocks built from zeroes. The number of points in $\mathcal{P}_F$ is $l/5+1$. Similarly to the case of the \textit{Zeromax-onemax} problem, there is only one solution that refers to each extreme $\mathcal{P}_F$ point, but there may be more solutions that refer to other parts of $\mathcal{P}_F$. Similarly to the single-objective optimisation, it is difficult to solve a problem built from deceptive blocks if a method is unable to obtain a linkage of appropriately high-quality \cite{MoGomeaSwarm,ltga}.

	\subsubsection{Leading Ones Trailing Zeros (LOTZ)}
	
	\textit{Leading ones trailing zeroes} is a classic benchmark in multi-objective optimisation. The first objective is the \textit{Leading Ones} function, while the second is \textit{Trailing Zeroes} function. They are defined as 
	
	\begin{equation}
	\label{eq:lotz}
	\mathit{f_{LO}(x)}=\sum_{i=1}^{l-1} \prod_{j=1}^{i} x_j; \mathit{f_{TZ}(x)}=\sum_{i=1}^{l-1} \prod_{j=i}^{l-1} (1-x_j).
	\end{equation}
	
	The leading ones function ($f_{LO}$) optimises the number of subsequent ones at the beginning of a genotype. The trailing zeroes optimises the number of subsequent zeroes at the end of it. The number of points in the Pareto-optimal front is $l+1$. In the case of LOTZ, each point in $\mathcal{P}_F$ refers to exactly one genotype.

	\subsubsection{Multi-objective Weighted MAXCUT}
	
	We employ the same multi-objective MAXCUT problem version as in \cite{MoGomeaSwarm,MoGomeaGecco} and use the same source code implementing the MAXCUT problem as in \cite{MoGomeaSwarm}. The problem instances have been generated using the approach proposed in \cite{maxcutOrigin}. The problem is defined as follows.\par
	
	Let $G=(V,E)$ be a weighted undirected graph, where $V=(v_0, v_1,\ldots,v_l)$ is a set of $l$ vertices and $E$ is the set of edges $(v_i,v_j)$. Each edge has an associated weight $w_{i,j}$. In the weighted MAXCUT problem, the objective is to find a maximum cut which is a partition of $l$ vertices into two disjoint subsets $A$ and $B$ ($A = V\ B$) such that the total weight of edges $(v_i,v_j)$ having $v_i \in A$ and $v_j \in B$ is maximised. We solve each MAXCUT instance for two different weight sets, making this problem bi-objective.\par

	The solution is encoded as a string of $l$ bits, where each variable $x_i$ corresponds to each vertex. If $x_i=0$ then $v_i \in A$ and $v_i \in B$ otherwise. The same solution encoding has been used in \cite{MoGomeaSwarm,MoGomeaGecco}. In the experiments, we consider problem instances for $l \in \{12, 25, 50, 100\}$. The Pareto-optimal front $\mathcal{P}_F$ is necessary to compute IGD. For $l \in \{12, 25\}$, the optimal Pareto front has been obtained by the enumeration method. For $l=50$ and larger $\mathcal{P}_F$, we use the reference sets proposed in \cite{MoGomeaGecco}. The instances and the reference Pareto front sets are the same as in \cite{MoGomeaSwarm,MoGomeaGecco}.

	\subsubsection{Multi-Objective Knapsack}
	
	In the multi-objective knapsack problem, we consider $l$ items and $m$ knapsacks. Each knapsack $k$ has capacity $c_k$ and each item $i$ is characterised by weight $w_{i,k}$ and profit $p_{i,k}$ corresponding to each knapsack $k$. Each item $i$ may be either selected and placed in every knapsack or not selected at all. Thus, the problem solution may be encoded as an $l$-bit binary string. If the total weight of the selected items does not violate the capacity constraint of any knapsacks, the solution is feasible. The objective is to maximise the profits of all knapsacks at the same time. The problem may be defined as
	\begin{equation}
	\label{eq:knapsack}
	\max_x (f_0(x), f_1(x),\ldots,f_{m-1}(x)),\\
	\end{equation}
	where $f_k(x) = \sum_{i=0}^{l-1}p_{i,k}x_i$ for $k=0,1,\ldots,m$ and subject to $\sum_{i=0}^{l-1} w_{i,k}x_i~\leq~c_k$.
	
	We use the same problem implementation as in \cite{MoGomeaSwarm}. Therefore, we also use the same mechanism to repair a solution that violates the constraints \cite{knapsackRepair}. The repair algorithm removes selected items one by one until all the constraints are satisfied. The items with the lowest profit/weight ratio are removed first.\par
	
	We employ the same bi-objective knapsack instances as in \cite{MoGomeaSwarm,knapsackRepair}. The considered number of items is $l \in \{100, 250, 500, 750\}$. For the instance of 750 items, we use pseudo-optimal $\mathcal{P}_F$ created by the combination of many Pareto fronts and employed in \cite{MoGomeaSwarm}. For the remaining instances, we use the optimal Pareto fronts reported in \cite{knapsackRepair}.

	\subsection{Main Results for Benchmarks}
	
	\begin{figure}
		\includegraphics[width=\linewidth]{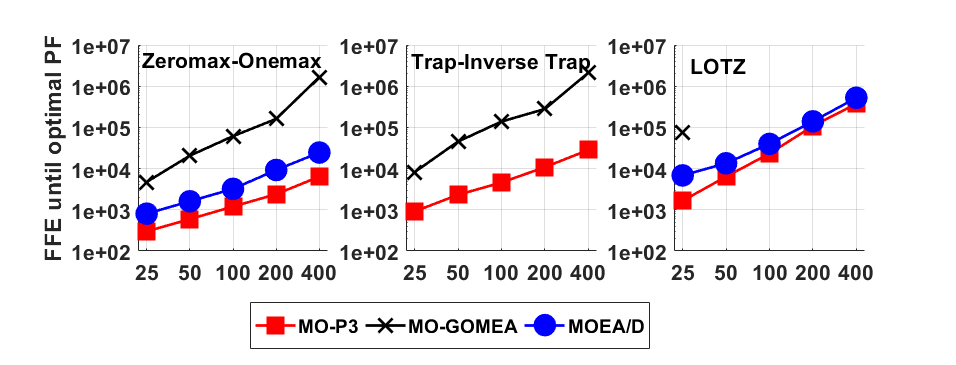}
		\caption{Scalability of MO-P3 and the competing methods for benchmark problems.}
		\label{fig:scalePerfsimpleBench}
	\end{figure}
	
	In Figure \ref{fig:scalePerfsimpleBench}, we show MO-P3, MO-GOMEA, and MOEA/D scalability on the \textit{Zeromax-Onemax}, \textit{Trap5-Inverse Trap5}, and \textit{LOTZ} problems. We present the median FFE necessary to find the optimal Pareto front. NSGA-II is excluded from the comparison because it has not found the optimal Pareto front in most of the runs for each problem. For these benchmarks, MO-P3 outperforms MO-GOMEA. This supremacy is caused by the following reason. For all three benchmarks, the linkage is the same for all Pareto front parts. For instance, for the \textit{Trap5-Inverse Trap5} problem, the corresponding bits always occupy the same blocks. This situation may favour MO-P3 because population clusterisation employed by MO-GOMEA does not guarantee any benefits. However, thanks to using linkage learning, MO-GOMEA is the only competing method that can successfully solve the \textit{Trap5-Inverse Trap5}. MOEA/D and NSGA-II were unable to find the optimal Pareto front even for a 25-bit problem version. Note that for the problems based on deceptive trap functions, the DSM-using methods (e.g., MO-GOMEA and MO-P3) are capable of finding the perfect problem decomposition in the early stages of the run \cite{linkageQuality}. This capability allows them to solve the problem.

	\begin{figure}
		\includegraphics[width=\linewidth]{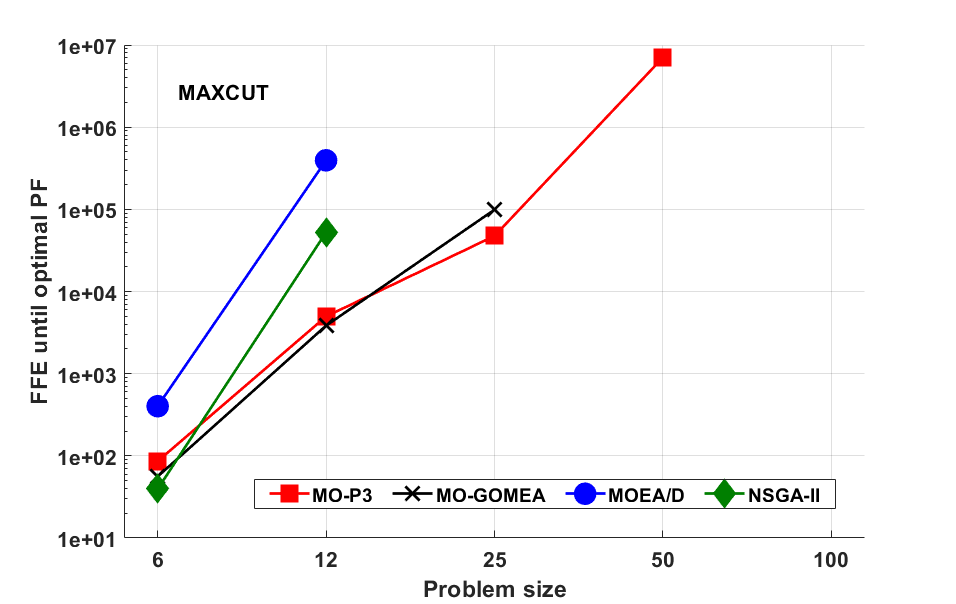}
		\caption{Scalability of MO-P3 and the competing methods for the Maxcut problem}
		\label{fig:scalabilityMaxcut}
	\end{figure}

	\begin{figure}
		\includegraphics[width=\linewidth]{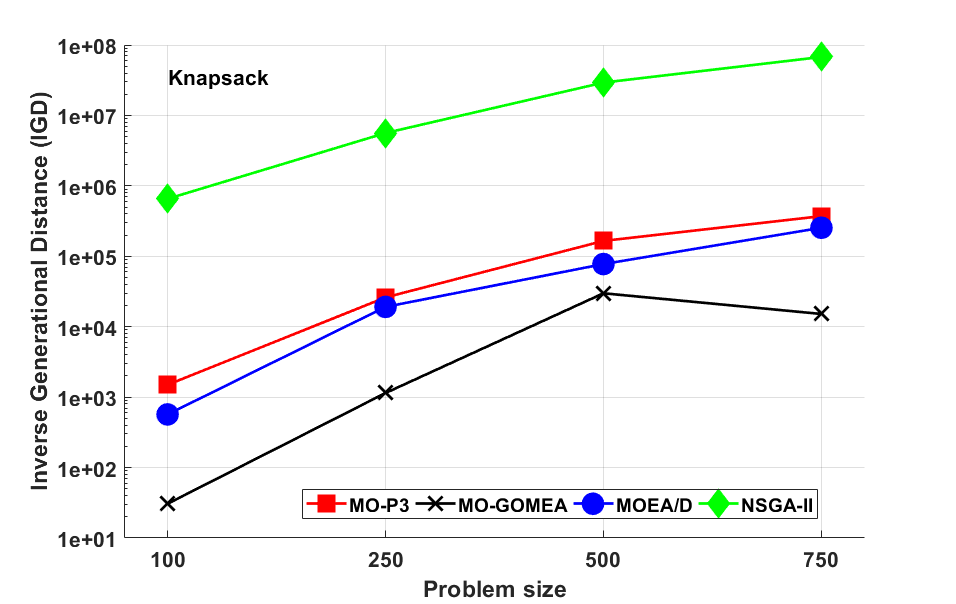}
		\caption{Scalability of MO-P3 and the competing methods for the Knapsack problem}
		\label{fig:scalabilityKnap}
	\end{figure}
	
	In Figure \ref{fig:scalabilityMaxcut}, we present the scalability on the MAXCUT problems. MO-P3 performs better than all the remaining considered methods including MO-GOMEA. It is the only method that can solve all the MAXCUT instances for $l \leq 50$. On the other hand, MO-GOMEA outperforms MO-P3 for the bi-objective Knapsack problem (Figure \ref{fig:scalabilityKnap}). For this problem, MO-P3 is also slightly outperformed by MOEA/D. Note that bi-objective knapsack problem is the only problem considered in this paper for which MO-P3 has been outperformed by any other method.

	\section{Results Discussion}
	In this paper, we have proposed a new multi-objective optimisation method based on the proposition of the Parameter-less Population Pyramid. MO-P3 uses a weighted vector to optimise new solutions added to the pyramid. Two different strategies have been considered: \textit{random} and \textit{smart}. The comparison based on a practical MOBCPP problem has shown that MO-P3-Random performs better. Such an observation may be surprising but is well explainable as the \textit{random} strategy does not bias MO-P3 towards any part of the Pareto front. The lack of bias allows MO-P3 to maintain a diverse linkage that seems to be the key to solve the considered practical problem. The importance of linkage is also supported by the comparison between MO-P3 and NSGA-II. For all the considered test cases except the smallest one, MO-P3 has outperformed NSGA-II as the IGD values for MO-P3 have been equal to $0$ in all the runs. It means that at least some of the points from the Pareto fronts proposed by NSGA-II have been dominated by the points from the Pareto fronts proposed by MO-P3.\par
	
	The comparison with MO-GOMEA on the base of MOBCPP instances shows that both the methods yield results of the same quality. However, for most of the test cases, MO-P3 requires significantly fewer FFEs to reach the best result. These results confirm the intuition presented in Section \ref{sec:mop3} that a P3-based method may be more suitable to solve practical problems than the LTGA-based ones. The high potential of MO-P3 application to practical problems is also confirmed by the analysis of the $FFE/ComputationTime$ ratio. It is significantly higher for MO-P3 than for MO-GOMEA and NSGA-II since MO-P3 does not spend the computation resources on the population clusterisation performed by MO-GOMEA and the computation of the crowding distance done by NSGA-II.\par

	The comparison made on the multi-objective benchmarks also indicates high effectiveness and high efficiency of MO-P3. For \textit{Zeromax-Onemax}, \textit{Trap5-Inverse Trap5} and \textit{LOTZ} problems, MO-P3 has found the optimal Pareto front in all runs, outperforming MO-GOMEA and NSGA-II. MO-GOMEA has been unable to find the optimal Pareto front for $l>25$. Moreover, for these three problems, MO-P3 has found the optimal Pareto fronts up to 100 times faster than MO-GOMEA. For the MAXCUT problem, MO-P3 has also outperformed the competing methods. Bi-objective Knapsack problem has been the only problem for which MO-GOMEA yielded Pareto fronts of better quality. The detailed analysis of MO-GOMEA and MO-P3 behavior for the Knapsack problem instances requires further investigation and is out of this paper's scope. However, it is possible that for the considered problem instances, MO-GOMEA can divide its population into such clusters that it obtains the linkage of high quality (the linkage that shows the true gene dependencies for the considered Pareto front parts).\par

	\section{Conclusions}
	In this paper, we have proposed MO-P3, a new method designed to solve a practical industrial multi-objective problem related to the process of production planning. The proposed method is adjusted to the problem with the appropriate solution encoding-decoding algorithm. Since solutions are represented by binary strings, MO-P3 has been compared with NSGA-II that is a typical baseline in multi-objective optimisation and MO-GOMEA that is an up-to-date method in solving multi-objective problems in discrete domains. Our proposition outperforms all competing methods for the set of considered practical problem instances.\par
	
	The experiments conducted on a set of typical benchmarks have confirmed both high-effectiveness and high-efficiency of MO-P3. The proposed method is not only capable of finding optimal or near-optimal Pareto fronts, but it is also the fastest in most of the cases. Additionally, it significantly outperforms both of the competing methods when the $FFE/ComputationTime$ ratio is taken into account. This positive feature is obtained thanks to the fact that MO-P3 does not require computationally costly operations like population clusterisation or crowding distance calculation.\par
	
	The main future work directions are as follows. The behaviour of MO-P3 for the Knapsack problem must be analysed to investigate the reason it performs worse than MO-GOMEA for this problem. The other future work objective is the development of MO-P3 itself to further improve its effectiveness and efficiency.

	\section*{Acknowledgement}
	We would like to thank Hoang Luong (Centrum Wiskunde \& Informatica) for giving us the access to the original MO-GOMEA source codes which has significantly speeded up the experiments preparation process and improved the paper quality.\par
	We also wish to express our gratitude to Marcin Komarnicki (Wroclaw University of Science and Technology) for helping us in organising the source codes and improving readability and composition of the paper.\par

	The authors acknowledge the support of the EU H2020 SAFIRE project (Ref. 723634).

	\bibliography{mop3}

\begin{thebibliography}{10}
\expandafter\ifx\csname url\endcsname\relax
  \def\url#1{\texttt{#1}}\fi
\expandafter\ifx\csname urlprefix\endcsname\relax\def\urlprefix{URL }\fi
\expandafter\ifx\csname href\endcsname\relax
  \def\href#1#2{#2} \def\path#1{#1}\fi

\bibitem{ltgaPopulationSizing}
P.~A. Bosman, N.~H. Luong, D.~Thierens, Expanding from discrete cartesian to
  permutation gene-pool optimal mixing evolutionary algorithms, in: Proceedings
  of the Genetic and Evolutionary Computation Conference 2016, GECCO '16, ACM,
  2016, pp. 637--644.

\bibitem{productionSchedulingIeee}
V.~{Santucci}, M.~{Baioletti}, A.~{Milani}, Algebraic differential evolution
  algorithm for the permutation flowshop scheduling problem with total flowtime
  criterion, IEEE Transactions on Evolutionary Computation 20~(5) (2016)
  682--694.

\bibitem{productionSchedulingMO}
J.~Deng, L.~Wang,
  \href{http://www.sciencedirect.com/science/article/pii/S2210650216300281}{A
  competitive memetic algorithm for multi-objective distributed permutation
  flow shop scheduling problem}, Swarm and Evolutionary Computation 32 (2017)
  121 -- 131.
\newblock \href {http://dx.doi.org/https://doi.org/10.1016/j.swevo.2016.06.002}
  {\path{doi:https://doi.org/10.1016/j.swevo.2016.06.002}}.
\newline\urlprefix\url{http://www.sciencedirect.com/science/article/pii/S2210650216300281}

\bibitem{productionMOrealNumber}
N.-Z. Lee, P.~Arcaini, S.~Ali, F.~Ishikawa,
  \href{https://doi.org/10.1145/3321707.3321755}{Stability analysis for safety
  of automotive multi-product lines: A search-based approach}, in: Proceedings
  of the Genetic and Evolutionary Computation Conference, GECCO ’19,
  Association for Computing Machinery, New York, NY, USA, 2019, p. 1241–1249.
\newblock \href {http://dx.doi.org/10.1145/3321707.3321755}
  {\path{doi:10.1145/3321707.3321755}}.
\newline\urlprefix\url{https://doi.org/10.1145/3321707.3321755}

\bibitem{productionSchedulingManyO}
A.~{Masood}, Y.~{Mei}, G.~{Chen}, M.~{Zhang}, Many-objective genetic
  programming for job-shop scheduling, in: 2016 IEEE Congress on Evolutionary
  Computation (CEC), 2016, pp. 209--216.

\bibitem{productionYorkRealNumbers}
P.~Dziurzanski, J.~Swan, L.~S. Indrusiak,
  \href{https://doi.org/10.1145/3205455.3205501}{Value-based manufacturing
  optimisation in serverless clouds for industry 4.0}, in: Proceedings of the
  Genetic and Evolutionary Computation Conference, GECCO ’18, Association for
  Computing Machinery, New York, NY, USA, 2018, p. 1222–1229.
\newblock \href {http://dx.doi.org/10.1145/3205455.3205501}
  {\path{doi:10.1145/3205455.3205501}}.
\newline\urlprefix\url{https://doi.org/10.1145/3205455.3205501}

\bibitem{productionDiscrete}
L.~F. {de Araújo Pessoa}, B.~{Hellingrath}, F.~B. {de Lima Neto}, Automatic
  generation of optimization algorithms for production lot-sizing problems, in:
  2019 IEEE Congress on Evolutionary Computation (CEC), 2019, pp. 1774--1781.

\bibitem{paintsOrig}
P.~Dziurzanski, S.~Zhao, J.~Swan, L.~S. Indrusiak, S.~Scholze, K.~Krone,
  Solving the multi-objective flexible job-shop scheduling problem with
  alternative recipes for a chemical production process, in: P.~Kaufmann, P.~A.
  Castillo (Eds.), Applications of Evolutionary Computation, Springer
  International Publishing, Cham, 2019, pp. 33--48.

\bibitem{Garey1990}
M.~R. Garey, D.~S. Johnson, Computers and Intractability; A Guide to the Theory
  of NP-Completeness, W. H. Freeman \& Co., New York, NY, USA, 1990.

\bibitem{MoGomeaGecco}
N.~H. Luong, H.~La~Poutr{\'e}, P.~A. Bosman,
  \href{http://doi.acm.org/10.1145/2576768.2598261}{Multi-objective gene-pool
  optimal mixing evolutionary algorithms}, in: Proceedings of the 2014 Annual
  Conference on Genetic and Evolutionary Computation, GECCO '14, ACM, New York,
  NY, USA, 2014, pp. 357--364.
\newblock \href {http://dx.doi.org/10.1145/2576768.2598261}
  {\path{doi:10.1145/2576768.2598261}}.
\newline\urlprefix\url{http://doi.acm.org/10.1145/2576768.2598261}

\bibitem{EliteArchive}
C.~A. Coello~Coello, G.~T. Pulido, E.~M. Montes,
  \href{https://doi.org/10.1007/1-84628-117-2_15}{Current and Future Research
  Trends in Evolutionary Multiobjective Optimization}, Springer London, London,
  2005, pp. 213--231.
\newblock \href {http://dx.doi.org/10.1007/1-84628-117-2_15}
  {\path{doi:10.1007/1-84628-117-2_15}}.
\newline\urlprefix\url{https://doi.org/10.1007/1-84628-117-2_15}

\bibitem{muppetsBaldwinEon}
M.~W. Przewozniczek, K.~Walkowiak, M.~Aibin,
  \href{https://doi.org/10.1016/j.asoc.2016.09.040}{The evolutionary cost of
  baldwin effect in the routing and spectrum allocation problem in elastic
  optical networks}, Appl. Soft Comput. 52~(C) (2017) 843--862.
\newblock \href {http://dx.doi.org/10.1016/j.asoc.2016.09.040}
  {\path{doi:10.1016/j.asoc.2016.09.040}}.
\newline\urlprefix\url{https://doi.org/10.1016/j.asoc.2016.09.040}

\bibitem{nsga2}
K.~{Deb}, A.~{Pratap}, S.~{Agarwal}, T.~{Meyarivan}, A fast and elitist
  multiobjective genetic algorithm: Nsga-ii, IEEE Transactions on Evolutionary
  Computation 6~(2) (2002) 182--197.
\newblock \href {http://dx.doi.org/10.1109/4235.996017}
  {\path{doi:10.1109/4235.996017}}.

\bibitem{moead}
Q.~{Zhang}, H.~{Li}, Moea/d: A multiobjective evolutionary algorithm based on
  decomposition, IEEE Transactions on Evolutionary Computation 11~(6) (2007)
  712--731.
\newblock \href {http://dx.doi.org/10.1109/TEVC.2007.892759}
  {\path{doi:10.1109/TEVC.2007.892759}}.

\bibitem{Zhou2011}
A.~Zhou, B.-Y. Qu, H.~Li, S.-Z. Zhao, P.~N. Suganthan, Q.~Zhang,
  \href{http://www.sciencedirect.com/science/article/pii/S2210650211000058}{Multiobjective
  evolutionary algorithms: A survey of the state of the art}, Swarm and
  Evolutionary Computation 1~(1) (2011) 32 -- 49.
\newblock \href {http://dx.doi.org/https://doi.org/10.1016/j.swevo.2011.03.001}
  {\path{doi:https://doi.org/10.1016/j.swevo.2011.03.001}}.
\newline\urlprefix\url{http://www.sciencedirect.com/science/article/pii/S2210650211000058}

\bibitem{Ma2014b}
X.~Ma, F.~Liu, Y.~Qi, M.~Gong, M.~Yin, L.~Li, L.~Jiao, J.~Wu,
  \href{https://doi.org/10.1016/j.neucom.2014.04.068}{Moea/d with
  opposition-based learning for multiobjective optimization problem},
  Neurocomput. 146~(C) (2014) 48–64.
\newblock \href {http://dx.doi.org/10.1016/j.neucom.2014.04.068}
  {\path{doi:10.1016/j.neucom.2014.04.068}}.
\newline\urlprefix\url{https://doi.org/10.1016/j.neucom.2014.04.068}

\bibitem{Ma2014}
X.~Ma, F.~Liu, Y.~Qi, L.~Li, L.~Jiao, M.~Liu, J.~Wu,
  \href{http://www.sciencedirect.com/science/article/pii/S0925231214006390}{Moea/d
  with baldwinian learning inspired by the regularity property of continuous
  multiobjective problem}, Neurocomputing 145 (2014) 336 -- 352.
\newblock \href
  {http://dx.doi.org/https://doi.org/10.1016/j.neucom.2014.05.025}
  {\path{doi:https://doi.org/10.1016/j.neucom.2014.05.025}}.
\newline\urlprefix\url{http://www.sciencedirect.com/science/article/pii/S0925231214006390}

\bibitem{Ma2016}
X.~Ma, F.~Liu, Y.~Qi, L.~Li, L.~Jiao, X.~Deng, X.~Wang, B.~Dong, Z.~Hou,
  Y.~Zhang, J.~Wu, \href{https://doi.org/10.1007/s00500-015-1789-z}{Moea/d with
  biased weight adjustment inspired by user preference and its application on
  multi-objective reservoir flood control problem}, Soft Comput. 20~(12) (2016)
  4999–5023.
\newblock \href {http://dx.doi.org/10.1007/s00500-015-1789-z}
  {\path{doi:10.1007/s00500-015-1789-z}}.
\newline\urlprefix\url{https://doi.org/10.1007/s00500-015-1789-z}

\bibitem{MoGomeaSwarm}
N.~H. Luong, H.~L. Poutré, P.~A. Bosman,
  \href{http://www.sciencedirect.com/science/article/pii/S2210650217304765}{Multi-objective
  gene-pool optimal mixing evolutionary algorithm with the interleaved
  multi-start scheme}, Swarm and Evolutionary Computation 40 (2018) 238 -- 254.
\newblock \href {http://dx.doi.org/https://doi.org/10.1016/j.swevo.2018.02.005}
  {\path{doi:https://doi.org/10.1016/j.swevo.2018.02.005}}.
\newline\urlprefix\url{http://www.sciencedirect.com/science/article/pii/S2210650217304765}

\bibitem{muppets}
H.~Kwasnicka, M.~Przewozniczek, Multi population pattern searching algorithm: A
  new evolutionary method based on the idea of messy genetic algorithm, IEEE
  Trans. Evolutionary Computation 15 (2011) 715--734.

\bibitem{3lo}
M.~W. Przewozniczek, M.~M. Komarnicki, Empirical linkage learning, IEEE Trans.
  Evolutionary Computation (2020) (in press).

\bibitem{ltga}
D.~Thierens, P.~A. Bosman,
  \href{http://doi.acm.org/10.1145/2463372.2463477}{Hierarchical problem
  solving with the linkage tree genetic algorithm}, in: Proceedings of the 15th
  Annual Conference on Genetic and Evolutionary Computation, GECCO '13, ACM,
  New York, NY, USA, 2013, pp. 877--884.
\newblock \href {http://dx.doi.org/10.1145/2463372.2463477}
  {\path{doi:10.1145/2463372.2463477}}.
\newline\urlprefix\url{http://doi.acm.org/10.1145/2463372.2463477}

\bibitem{subpopInitLL}
M.~W. Przewoźniczek,
  \href{http://www.sciencedirect.com/science/article/pii/S002002552030102X}{Subpopulation
  initialization driven by linkage learning for dealing with the
  long-way-to-stuck effect}, Information Sciences 521 (2020) 62 -- 80.
\newblock \href {http://dx.doi.org/https://doi.org/10.1016/j.ins.2020.02.027}
  {\path{doi:https://doi.org/10.1016/j.ins.2020.02.027}}.
\newline\urlprefix\url{http://www.sciencedirect.com/science/article/pii/S002002552030102X}

\bibitem{mupMemo}
M.~W. Przewoźniczek, R.~Goścień,
  \href{http://www.sciencedirect.com/science/article/pii/S1568494619303722}{Universal
  strategy of dynamic subpopulation number management in practical network
  optimization problems}, Applied Soft Computing 82 (2019) 105592.
\newblock \href {http://dx.doi.org/https://doi.org/10.1016/j.asoc.2019.105592}
  {\path{doi:https://doi.org/10.1016/j.asoc.2019.105592}}.
\newline\urlprefix\url{http://www.sciencedirect.com/science/article/pii/S1568494619303722}

\bibitem{mohBOA}
M.~Pelikan, K.~Sastry, D.~E. Goldberg,
  \href{http://doi.acm.org/10.1145/1068009.1068122}{Multiobjective hboa,
  clustering, and scalability}, in: Proceedings of the 7th Annual Conference on
  Genetic and Evolutionary Computation, GECCO '05, ACM, New York, NY, USA,
  2005, pp. 663--670.
\newblock \href {http://dx.doi.org/10.1145/1068009.1068122}
  {\path{doi:10.1145/1068009.1068122}}.
\newline\urlprefix\url{http://doi.acm.org/10.1145/1068009.1068122}

\bibitem{ltgaGomeaNaming}
P.~A. Bosman, D.~Thierens, More concise and robust linkage learning by
  filtering and combining linkage hierarchies, in: Proceedings of the 15th
  Annual Conference on Genetic and Evolutionary Computation, GECCO '13, ACM,
  New York, NY, USA, 2013, pp. 359--366.

\bibitem{P3Original}
B.~W. Goldman, W.~F. Punch,
  \href{http://doi.acm.org/10.1145/2576768.2598350}{Parameter-less population
  pyramid}, in: Proceedings of the 2014 Annual Conference on Genetic and
  Evolutionary Computation, GECCO '14, ACM, New York, NY, USA, 2014, pp.
  785--792.
\newblock \href {http://dx.doi.org/10.1145/2576768.2598350}
  {\path{doi:10.1145/2576768.2598350}}.
\newline\urlprefix\url{http://doi.acm.org/10.1145/2576768.2598350}

\bibitem{dsmga2}
S.-H. Hsu, T.-L. Yu,
  \href{http://doi.acm.org/10.1145/2739480.2754737}{Optimization by pairwise
  linkage detection, incremental linkage set, and restricted / back mixing:
  Dsmga-ii}, in: Proceedings of the 2015 Annual Conference on Genetic and
  Evolutionary Computation, GECCO '15, ACM, New York, NY, USA, 2015, pp.
  519--526.
\newblock \href {http://dx.doi.org/10.1145/2739480.2754737}
  {\path{doi:10.1145/2739480.2754737}}.
\newline\urlprefix\url{http://doi.acm.org/10.1145/2739480.2754737}

\bibitem{linkageQuality}
M.~W. Przewozniczek, B.~Frej, M.~M. Komarnicki, On measuring and improving the
  quality of linkage learning in modern evolutionary algorithms applied to
  solve partially additively separable problems, in: Proceedings of the 2020
  Annual Conference on Genetic and Evolutionary Computation, GECCO '20, 2020,
  p. (in press).

\bibitem{linkLearningDetermined}
D.~Thierens, P.~Bosman,
  \href{https://doi.org/10.1145/2330163.2330205}{Predetermined versus learned
  linkage models}, in: Proceedings of the 14th Annual Conference on Genetic and
  Evolutionary Computation, GECCO ’12, Association for Computing Machinery,
  New York, NY, USA, 2012, p. 289–296.
\newblock \href {http://dx.doi.org/10.1145/2330163.2330205}
  {\path{doi:10.1145/2330163.2330205}}.
\newline\urlprefix\url{https://doi.org/10.1145/2330163.2330205}

\bibitem{watsonHiff}
R.~A. Watson, J.~B. Pollack, Hierarchically consistent test problems for
  genetic algorithms, in: Proceedings of the 1999 Congress on Evolutionary
  Computation-CEC99 (Cat. No. 99TH8406), Vol.~2, 1999, pp. 1406--1413 Vol. 2.
\newblock \href {http://dx.doi.org/10.1109/CEC.1999.782647}
  {\path{doi:10.1109/CEC.1999.782647}}.

\bibitem{watsonHiffPPSNfirst}
J.~B.~P. R.~A.~Watson, G. S.~Hornby, Hierarchical building-block problems for
  ga evaluation, in: Proceedings of the 1998 International Conference on
  Parallel Problem Solving from Nature, Vol.~2, 1998, pp. 97--106.

\bibitem{linkageLearningIsBad}
J.~P. Martins, C.~M. Fonseca, A.~C. Delbem,
  \href{http://www.sciencedirect.com/science/article/pii/S0925231214008807}{On
  the performance of linkage-tree genetic algorithms for the multidimensional
  knapsack problem}, Neurocomputing 146 (2014) 17 -- 29, bridging Machine
  learning and Evolutionary Computation (BMLEC) Computational Collective
  Intelligence.
\newblock \href
  {http://dx.doi.org/https://doi.org/10.1016/j.neucom.2014.04.069}
  {\path{doi:https://doi.org/10.1016/j.neucom.2014.04.069}}.
\newline\urlprefix\url{http://www.sciencedirect.com/science/article/pii/S0925231214008807}

\bibitem{fP3}
M.~M. Komarnicki, M.~W. Przewozniczek, Parameter-less population pyramid with
  feedback, in: Proceedings of the Genetic and Evolutionary Computation
  Conference Companion, GECCO '17, ACM, 2017, pp. 109--110.

\bibitem{afP3}
A.~M. Zielinski, M.~M. Komarnicki, M.~W. Przewozniczek,
  \href{http://doi.acm.org/10.1145/3319619.3322052}{Parameter-less population
  pyramid with automatic feedback}, in: Proceedings of the Genetic and
  Evolutionary Computation Conference Companion, GECCO '19, ACM, New York, NY,
  USA, 2019, pp. 312--313.
\newblock \href {http://dx.doi.org/10.1145/3319619.3322052}
  {\path{doi:10.1145/3319619.3322052}}.
\newline\urlprefix\url{http://doi.acm.org/10.1145/3319619.3322052}

\bibitem{ieeeSurvey}
X.~{Ma}, X.~{Li}, Q.~{Zhang}, K.~{Tang}, Z.~{Liang}, W.~{Xie}, Z.~{Zhu}, A
  survey on cooperative co-evolutionary algorithms, IEEE Transactions on
  Evolutionary Computation 23~(3) (2019) 421--441.

\bibitem{llClassification}
Y.-p. Chen, T.-L. Yu, K.~Sastry, D.~E. Goldberg, A survey of linkage learning
  techniques in genetic and evolutionary algorithms, Illinois Genetic
  Algorithms Library, Tech. Rep. (2007).

\bibitem{muppetsActive}
M.~Przewozniczek, \href{http://dx.doi.org/10.1016/j.ins.2016.02.048}{Active
  multi-population pattern searching algorithm for flow optimization in
  computer networks - the novel coevolution schema combined with linkage
  learning}, Inf. Sci. 355~(C) (2016) 15--36.
\newblock \href {http://dx.doi.org/10.1016/j.ins.2016.02.048}
  {\path{doi:10.1016/j.ins.2016.02.048}}.
\newline\urlprefix\url{http://dx.doi.org/10.1016/j.ins.2016.02.048}

\bibitem{DSMorig}
T.-L. Yu, D.~E. Goldberg, K.~Sastry, C.~F. Lima, M.~Pelikan, Dependency
  structure matrix, genetic algorithms, and effective recombination,
  Evolutionary Computation 17 (2009) 595--626.

\bibitem{omidvar}
M.~N. Omidvar, X.~Li, Y.~Mei, X.~Yao, Cooperative co-evolution with
  differential grouping for large scale optimization, IEEE Transactions on
  Evolutionary Computation 18~(3) (2014) 378--393.
\newblock \href {http://dx.doi.org/10.1109/TEVC.2013.2281543}
  {\path{doi:10.1109/TEVC.2013.2281543}}.

\bibitem{linkageRandom}
Z.~Yang, K.~Tang, X.~Yao,
  \href{http://www.sciencedirect.com/science/article/pii/S002002550800073X}{Large
  scale evolutionary optimization using cooperative coevolution}, Information
  Sciences 178~(15) (2008) 2985 -- 2999, nature Inspired Problem-Solving.
\newblock \href {http://dx.doi.org/https://doi.org/10.1016/j.ins.2008.02.017}
  {\path{doi:https://doi.org/10.1016/j.ins.2008.02.017}}.
\newline\urlprefix\url{http://www.sciencedirect.com/science/article/pii/S002002550800073X}

\bibitem{linkageClassificationCC}
X.~Li, Q.~Zhang, K.~Tang, Z.~Liang, W.~Xie, Z.~Zhu, A survey on cooperative
  co-evolutionary algorithms, IEEE Transactions on Evolutionary Computation PP
  (2018) 1--1.
\newblock \href {http://dx.doi.org/10.1109/TEVC.2018.2868770}
  {\path{doi:10.1109/TEVC.2018.2868770}}.

\bibitem{mutualInformation}
S.~Kullback, R.~A. Leibler, On information and sufficiency, Ann. Math. Statist.
  22~(1) (1951) 79--86.

\bibitem{psDSMGA2}
M.~M. Komarnicki, M.~W. Przewozniczek,
  \href{http://doi.acm.org/10.1145/3319619.3322080}{Parameter-less,
  population-sizing dsmga-ii}, in: Proceedings of the Genetic and Evolutionary
  Computation Conference Companion, GECCO '19, ACM, New York, NY, USA, 2019,
  pp. 289--290.
\newblock \href {http://dx.doi.org/10.1145/3319619.3322080}
  {\path{doi:10.1145/3319619.3322080}}.
\newline\urlprefix\url{http://doi.acm.org/10.1145/3319619.3322080}

\bibitem{dsmga2e}
P.-L. Chen, C.-J. Peng, C.-Y. Lu, T.-L. Yu,
  \href{http://doi.acm.org/10.1145/3071178.3071236}{Two-edge graphical linkage
  model for dsmga-ii}, in: Proceedings of the Genetic and Evolutionary
  Computation Conference, GECCO '17, ACM, New York, NY, USA, 2017, pp.
  745--752.
\newblock \href {http://dx.doi.org/10.1145/3071178.3071236}
  {\path{doi:10.1145/3071178.3071236}}.
\newline\urlprefix\url{http://doi.acm.org/10.1145/3071178.3071236}

\bibitem{steinerTrees}
M.~W. Przewozniczek, K.~Walkowiak, A.~Sen, M.~Komarnicki, P.~Lechowicz,
  \href{https://doi.org/10.1016/j.ins.2018.11.015}{The transformation of the
  k-shortest steiner trees search problem into binary dynamic problem for
  effective evolutionary methods application}, Inf. Sci. 479 (2019) 1--19.
\newblock \href {http://dx.doi.org/10.1016/j.ins.2018.11.015}
  {\path{doi:10.1016/j.ins.2018.11.015}}.
\newline\urlprefix\url{https://doi.org/10.1016/j.ins.2018.11.015}

\bibitem{OverlappingSimon}
H.~A. Simon, The sciences of the artificial, Cambridge, MA. MIT Press.

\bibitem{decFunc}
K.~Deb, D.~E. Goldberg, Sufficient conditions for deceptive and easy binary
  functions, Ann. Math. Artif. Intell. 10~(4) (1993) 385--408.

\bibitem{grayWhitley}
L.~D. {Whitley}, F.~{Chicano}, B.~W. {Goldman}, Gray box optimization for mk
  landscapes (nk landscapes and max-ksat), Evolutionary Computation 24~(3)
  (2016) 491--519.
\newblock \href {http://dx.doi.org/10.1162/EVCO_a_00184}
  {\path{doi:10.1162/EVCO_a_00184}}.

\bibitem{dsmga2PopulationSizing}
M.~M. Komarnicki, M.~W. Przewozniczek, Parameter-less, population-sizing
  dsmga-ii, in: Proceedings of the Genetic and Evolutionary Computation
  Conference Companion, GECCO '19, ACM, 2019 (in press).

\bibitem{frontQualityMeasurement}
M.~{Laumanns}, L.~{Thiele}, K.~{Deb}, E.~{Zitzler}, Combining convergence and
  diversity in evolutionary multiobjective optimization, Evolutionary
  Computation 10~(3) (2002) 263--282.
\newblock \href {http://dx.doi.org/10.1162/106365602760234108}
  {\path{doi:10.1162/106365602760234108}}.

\bibitem{spea2}
E.~Zitzler, M.~Laumanns, L.~Thiele, Spea2: Improving the strength pareto
  evolutionary algorithm for multiobjective optimization, Vol. 3242, 2001, pp.
  95--–100.

\bibitem{clusterizationBosman}
P.~A. Bosman, \href{http://doi.acm.org/10.1145/1830483.1830549}{The anticipated
  mean shift and cluster registration in mixture-based edas for multi-objective
  optimization}, in: Proceedings of the 12th Annual Conference on Genetic and
  Evolutionary Computation, GECCO '10, ACM, New York, NY, USA, 2010, pp.
  351--358.
\newblock \href {http://dx.doi.org/10.1145/1830483.1830549}
  {\path{doi:10.1145/1830483.1830549}}.
\newline\urlprefix\url{http://doi.acm.org/10.1145/1830483.1830549}

\bibitem{ElitistArchive2}
N.~H. Luong, H.~La~Poutr{\'e}, P.~A. Bosman,
  \href{http://doi.acm.org/10.1145/2576768.2598261}{Multi-objective gene-pool
  optimal mixing evolutionary algorithms}, in: Proceedings of the 2014 Annual
  Conference on Genetic and Evolutionary Computation, GECCO '14, ACM, New York,
  NY, USA, 2014, pp. 357--364.
\newblock \href {http://dx.doi.org/10.1145/2576768.2598261}
  {\path{doi:10.1145/2576768.2598261}}.
\newline\urlprefix\url{http://doi.acm.org/10.1145/2576768.2598261}

\bibitem{Ma2014c}
X.~Ma, Y.~Qi, L.~Li, F.~Liu, L.~Jiao, J.~Wu, Moea/d with uniform decomposition
  measurement for many-objective problems, Soft Computing 18 (2014) 2541--2564.
\newblock \href {http://dx.doi.org/10.1007/s00500-014-1234-8}
  {\path{doi:10.1007/s00500-014-1234-8}}.

\bibitem{Ishibuchi1998}
H.~Ishibuchi, T.~Murata, A multi-objective genetic local search algorithm and
  its application to flowshop scheduling, Trans. Sys. Man Cyber Part C 28~(3)
  (1998) 392--403.

\bibitem{Li2009}
L.~Li, J.-Z. Huo, Multi-objective flexible job-shop scheduling problem in steel
  tubes production, Systems Engineering - Theory \& Practice 29~(8) (2009)
  117--126.

\bibitem{Huynh2018}
N.-T. Huynh, C.-F. Chien, A hybrid multi-subpopulation genetic algorithm for
  textile batch dyeing scheduling and an empirical study, Computers \&
  Industrial Engineering 125 (2018) 615--627.

\bibitem{Gen2014}
M.~Gen, L.~Lin, Multiobjective evolutionary algorithm for manufacturing
  scheduling problems: state-of-the-art survey, Journal of Intelligent
  Manufacturing 25~(5) (2014) 849--866.

\bibitem{moSurvey}
A.~Zhou, B.-Y. Qu, H.~Li, S.-Z. Zhao, P.~N. Suganthan, Q.~Zhang,
  \href{http://www.sciencedirect.com/science/article/pii/S2210650211000058}{Multiobjective
  evolutionary algorithms: A survey of the state of the art}, Swarm and
  Evolutionary Computation 1~(1) (2011) 32 -- 49.
\newblock \href {http://dx.doi.org/https://doi.org/10.1016/j.swevo.2011.03.001}
  {\path{doi:https://doi.org/10.1016/j.swevo.2011.03.001}}.
\newline\urlprefix\url{http://www.sciencedirect.com/science/article/pii/S2210650211000058}

\bibitem{linkageLearningIsBad2}
J.~P. Martins, A.~C. Delbem,
  \href{http://www.sciencedirect.com/science/article/pii/S2210650218300269}{Reproductive
  bias, linkage learning and diversity preservation in bi-objective
  evolutionary optimization}, Swarm and Evolutionary Computation 48 (2019) 145
  -- 155.
\newblock \href {http://dx.doi.org/https://doi.org/10.1016/j.swevo.2019.04.005}
  {\path{doi:https://doi.org/10.1016/j.swevo.2019.04.005}}.
\newline\urlprefix\url{http://www.sciencedirect.com/science/article/pii/S2210650218300269}

\bibitem{fitnessCaching}
M.~W. Przewozniczek, M.~M. Komarnicki, The influence of fitness caching on
  modern evolutionary methods and fair computation load measurement, in:
  Proceedings of the Genetic and Evolutionary Computation Conference Companion,
  GECCO '18, ACM, 2018, pp. 241--242.

\bibitem{PEACh}
M.~W. Przewo\'zniczek, Problem encoding allowing cheap fitness computation of
  mutated individuals, in: 2017 IEEE Congress on Evolutionary Computation
  (CEC), 2017, pp. 308--316.
\newblock \href {http://dx.doi.org/10.1109/CEC.2017.7969328}
  {\path{doi:10.1109/CEC.2017.7969328}}.

\bibitem{PEAChWPlfl}
M.~Przewozniczek, M.~Komarnicki, The practical use of problem encoding allowing
  cheap fitness computation of mutated individuals, in: 2018 Federated
  Conference on Computer Science and Information Systems (FedCSIS), 2018, pp.
  57--65.

\bibitem{maxcutOrigin}
P.~A. Bosman, D.~Thierens,
  \href{http://doi.acm.org/10.1145/2463372.2463420}{More concise and robust
  linkage learning by filtering and combining linkage hierarchies}, in:
  Proceedings of the 15th Annual Conference on Genetic and Evolutionary
  Computation, GECCO '13, ACM, New York, NY, USA, 2013, pp. 359--366.
\newblock \href {http://dx.doi.org/10.1145/2463372.2463420}
  {\path{doi:10.1145/2463372.2463420}}.
\newline\urlprefix\url{http://doi.acm.org/10.1145/2463372.2463420}

\bibitem{knapsackRepair}
E.~{Zitzler}, L.~{Thiele}, Multiobjective evolutionary algorithms: a
  comparative case study and the strength pareto approach, IEEE Transactions on
  Evolutionary Computation 3~(4) (1999) 257--271.
\newblock \href {http://dx.doi.org/10.1109/4235.797969}
  {\path{doi:10.1109/4235.797969}}.

\end{thebibliography}
	
\end{document}